\documentclass[10pt,journal,compsoc]{IEEEtran}

\ifCLASSOPTIONcompsoc
  \usepackage[nocompress]{cite}
\else
  \usepackage{cite}
\fi

\ifCLASSINFOpdf
\else
\fi
\usepackage{url}
\usepackage{hyperref}

\usepackage{graphicx}
\usepackage{comment}
\usepackage{amsmath,amssymb} % define this before the line numbering.
\usepackage{color}

\usepackage{cite}
\usepackage{dsfont}
\usepackage{nccmath}
\usepackage{textcomp}
\usepackage{arydshln}
\usepackage{multirow}
\usepackage{floatrow}

\graphicspath{{./figures/}}

\begin{document}
\title{CIE XYZ Net: Unprocessing Images for Low-Level Computer Vision Tasks}

\author{Mahmoud~Afifi
		%,~\IEEEmembership{Member,~IEEE,}
		\quad
        Abdelrahman~Abdelhamed
        %,~\IEEEmembership{Fellow,~OSA,}
        \quad
        Abdullah~Abuolaim\\
        \quad
        Abhijith~Punnappurath
        \quad
        ~Michael~S.~Brown
        %,~\IEEEmembership{Life~Fellow,~IEEE}% <-this % stops a space
\IEEEcompsocitemizethanks{%\IEEEcompsocthanksitem M. Shell was with the Department of Electrical and Computer Engineering, Georgia Institute of Technology, Atlanta, GA, 30332.\protect\\
%E-mail: see http://www.michaelshell.org/contact.html
%\IEEEcompsocthanksitem J. Doe and J. Doe are with Anonymous University.%
\IEEEcompsocthanksitem All authors are with the Department of Electrical Engineering and Computer Science, Lassonde School of Engineering, York University, Toronto, ON, Canada. \protect\\
E-mails: \protect\href{mailto:mafifi@eecs.yorku.ca; kamel@eecs.yorku.ca; abuolaim@eecs.yorku.ca; pabhijith@eecs.yorku.ca; mbrown@eecs.yorku.ca}{\protect\{mafifi, kamel, abuolaim, pabhijith, mbrown\protect\}@eecs.yorku.ca}.
}% <-this % stops an unwanted space
\thanks{}%
}

\markboth{}%
{Afifi \MakeLowercase{\textit{et al.}}: CIE XYZ Net: Unprocessing Images for Low-Level Computer Vision Tasks}

\IEEEtitleabstractindextext{%
\begin{abstract}
Cameras currently allow access to two image states: (i) a minimally processed linear raw-RGB image state (i.e., raw sensor data) or (ii) a highly-processed nonlinear image state (e.g., sRGB).  There are many computer vision tasks that work best with a linear image state, such as image deblurring and image dehazing.  Unfortunately, the vast majority of images are saved in the nonlinear image state.  Because of this, a number of methods have been proposed to ``unprocess'' nonlinear images back to a raw-RGB state.  However, existing unprocessing methods have a drawback because raw-RGB images are sensor-specific.  As a result, it is necessary to know which camera produced the sRGB output and use a method or network tailored for that sensor to properly unprocess it.  This paper addresses this limitation by exploiting another camera image state that is not available as an output, but it is available inside the camera pipeline.  In particular, cameras apply a colorimetric conversion step to convert the raw-RGB image to a device-independent space based on the CIE XYZ color space before they apply the nonlinear photo-finishing.   Leveraging this canonical image state, we propose a deep learning framework, CIE XYZ Net, that can unprocess a nonlinear image back to the canonical CIE XYZ image.   This image can then be processed by any low-level computer vision operator and re-rendered back to the nonlinear image.  We demonstrate the usefulness of the CIE XYZ Net on several low-level vision tasks and show significant gains that can be obtained by this processing framework. Code and dataset are publicly available at \url{https://github.com/mahmoudnafifi/CIE_XYZ_NET}.
\end{abstract}

}

\maketitle

\IEEEdisplaynontitleabstractindextext
\IEEEpeerreviewmaketitle

\IEEEraisesectionheading{\section{Introduction}\label{sec:intro}}
\IEEEPARstart{An} image signal processor (ISP) onboard a camera processes the initial captured sensor image in a pipeline fashion, with routines being applied one after the other.  The ISP used by consumer cameras performs operations as two distinct stages.  First, a ``front-end'' stage applies linear operations, such as white
balance and color adaptation, to convert the sensor-specific raw-RGB image to a device-independent color
space (e.g., CIE XYZ or its wide-gamut representation, ProPhoto)~\cite{karaimer2016software}. The image states associated with the front-end process are called a \textit{scene-referred} image because the image remains related directly to initial recorded sensor values related to the physical scene. Next, a ``photo-finishing'' stage is performed that applies nonlinear steps and local operators to produce a visually
pleasing photograph.  For example, selective color manipulation is often applied to enhance skin tone or make the overall colors more vivid, while local tone manipulation increases local contrast within the image.    After the photo-finishing stage, the image is encoded in an output color space (e.g., sRGB, AdobeRGB, or Display P3).  The image states associated with the photo-finishing process are referred to as \textit{display-referred} as they are encoded for visual display. Cameras currently allow access only to either the
minimally processed scene-referred image state (i.e., raw-RGB image) or the final display-referred image state (e.g., sRGB, AdobeRGB, or Display P3).  Unfortunately, these two image states are not ideal for low-level computer vision tasks.

\begin{figure}[t]
\centering
\includegraphics[width=\linewidth]{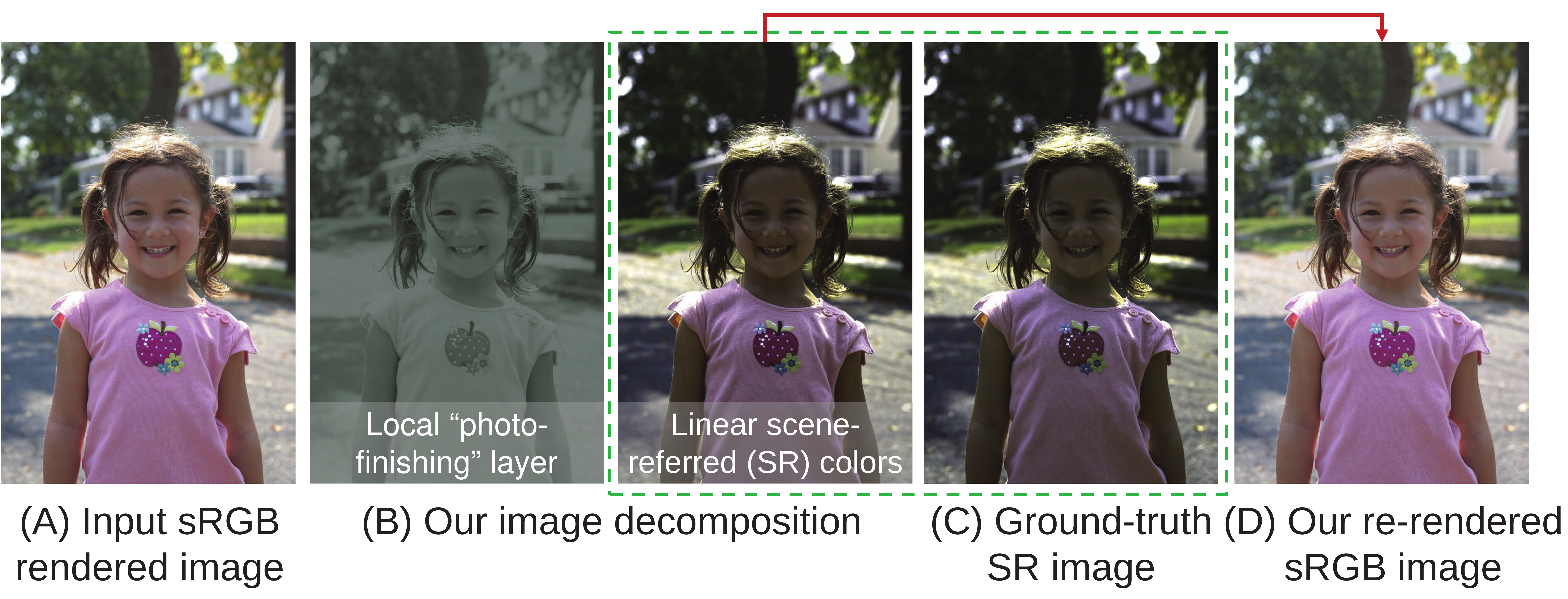}
\caption{We propose a cycle framework that can unprocess sRGB images back to the linear CIE XYZ color space and re-render the CIE XYZ images into the nonlinear sRGB color space. (A) The input camera-rendered sRGB image. (B) Our image decomposition (left: residual photo-finishing layer, right: scene-referred CIE XYZ reconstruction). (C) The ground-truth scene-referred CIE XYZ image. (D) Our re-rendering result from the reconstructed CIE XYZ image. To aid visualization, CIE XYZ images are scaled by a factor of two. Input image is taken from the MIT-Adobe FiveK dataset \cite{fivek}.}
\label{fig:teaser}
\end{figure}

The raw-RGB image state preserves the linear relationship of incident scene radiance. This linear image formation makes raw-RGB images suitable for a wide range of low-level computer vision tasks, such as image deblurring, image dehazing, image denoising, and various types of image enhancement~\cite{tai2013nonlinear, nguyen2016raw, Barron2019, zamir2020cycleisp}.  However, the drawback of raw-RGB is that the physical color filter arrays that make up the sensor's Bayer pattern are sensor-specific.  This means raw-RGB values captured of the same scene but with different sensors are significantly different~\cite{nguyen2014raw}.  This often requires learning-based methods to be trained per sensor or camera make and model (e.g.,~\cite{diamond2017dirty, nam2017modelling, hu2017fc4, Barron2019, afifi2019sensor}).

The more common display-referred image state (in this paper, assumed to be in the sRGB color space) also has drawbacks.  While this image state is the most widely used and is suitable for display,  cameras apply their own proprietary photo-finishing to enhance the visual quality of the image.  This means images captured of the same scene but using different camera models (and sometimes the same camera but with different settings) will produce images that have significantly different sRGB values~\cite{kim2012new, karaimer2016software, nguyen2016raw}.

As previously discussed, the front-end processor of a typical camera ISP performs a colorimetric conversion to map the raw-RGB image to a standard perceptual colorspace---namely, CIE 1931 XYZ~\cite{karaimer2016software}.  While there exists no formal image encoding for this image state, it is possible to convert existing raw-RGB images stored in digital negative (DNG) format to this intermediate state by applying a software camera ISP (e.g.,~\cite{abdelhamed2018high, karaimer2016software}). This provides a mechanism to standardize all images into a canonical linear scene-referred image state and is the impetus of our work.

%\vspace{2mm}
\noindent{\textbf{Contribution}}~We propose a method to decompose non-linear sRGB images into two parts: 1) a canonical linear scene-referred image state in the CIE XYZ color space and 2) a residual image layer that resembles additional non-linear and local photo-finishing operations. Through such decomposition strategy, we learn a model that can accurately map back and forth between non-linear sRGB and linear CIE XYZ images. An example is shown in Fig.~\ref{fig:teaser}. Unlike raw-RGB, the CIE XYZ color space is {\it device-independent}, and as a result, helps with model generalization. Furthermore, CIE XYZ images can be encoded as standard three-channel images that can be easily handled by existing computer vision frameworks. We show that our proposed model maps images back to the CIE XYZ color space more accurately compared to alternative approaches.  In addition, we perform extensive experiments on tasks such as image denoising, deblurring, and defocus estimation, to show that employing our proposed CIE XYZ model provides the performance boost anticipated from using linear images. Finally, we also show that using our decomposed image layers (CIE XYZ and a residual layer), our model can be used to perform various image enhancement and photo-finishing tasks.

\section{Related Work}\label{sec:related}

In this section, we first discuss the camera imaging pipeline that is necessary in order to understand how and why we access the camera image once converted to the CIE XYZ color space. We then review various methods proposed for linearization of camera-rendered images.

\subsection{Camera Imaging Pipeline} \label{subsec:related-pipe}

\begin{figure}[t]
\centering
\includegraphics[width=\linewidth]{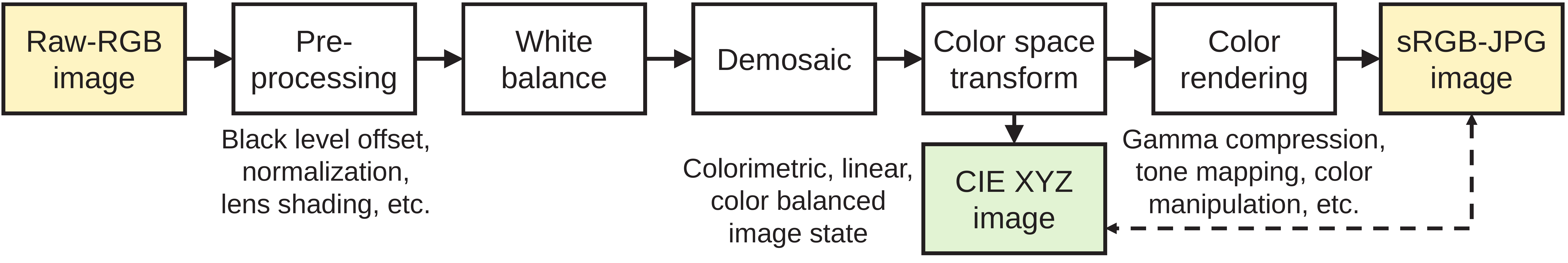}
\caption{A simplified depiction of a camera imaging pipeline, adapted from~\cite{gow2007comprehensive, liu2007automatic, hasinoff2010noise, karaimer2016software}. Our method allows mapping back and forth between the common nonlinear sRGB image and to the colorimetric, linear, color-balanced CIE XYZ image state for computer vision tasks.}
\label{fig:pipeline}
\end{figure}

Images captured by a camera undergo a sequence of processes in order to transform an initial image obtained by the raw sensor data (raw-RGB image) into a visually perceivable color space (e.g., sRGB) and a suitable format for storage and display (e.g., JPEG)~\cite{gow2007comprehensive, liu2007automatic, hasinoff2010noise, karaimer2016software}. A simplified depiction of the imaging pipeline is shown in Fig.~\ref{fig:pipeline}.  The pipeline processes can be divided into two main stages. The first stage is colorimetric processing  that transforms the image into a linear color space (e.g., CIE XYZ or ProPhoto) that preserves the direct mapping between the recorded scene values and the image. The second stage in the camera pipeline is the camera-rendering or photo-finishing stage that applies nonlinear transformations (e.g., gamma compression), selective color manipulation, and locally varying processing (e.g., local tone mapping) that break the relation between the image and the scene. Using images in the linear stages of the imaging pipeline has been shown to be more effective for image restoration and enhancement tasks~\cite{karaimer2016software,nguyen2016raw,Barron2019}. However, due to the complexity and proprietary nature of imaging pipelines on-board cameras, it is hard to obtain colorimetric images from cameras without the effort of saving raw sensor data or inverting the imaging pipeline stages~\cite{nguyen2016raw,Barron2019}.

\subsection{Camera-Rendered Image Linearization} \label{subsec:related-lin}

% map to raw
To obtain a linear image from its camera-rendered version, we need to reverse the nonlinear camera-rendering stage in the pipeline. Many methods have been proposed to model a parametric relationship that maps from the camera-rendered image (i.e., sRGB image) back to its raw-RGB version (e.g.,~\cite{nguyen2016raw}). However, raw-RGB space is camera-dependent and requires having a separate model per camera.
% standard linearization
Other approaches involve simple linearization by inverting the global tone mapping and the gamma compression followed by applying a linearization matrix to obtain a linear sRGB or CIE XYZ image~\cite{Barron2019}. Such approaches are too simple and do not account for the local processing or dynamic range adjustments.
% our approach
Unlike prior approaches, instead of trying only to obtain a linear image, our approach is to decompose the nonlinear image into globally processed and locally processed layers. The locally processed layer represents local color processing, such as local tone mapping. Then, we learn a global mapping from the globally processed image to the linear image.
% radiometric calibration
Another line of research targeting the problem of image linearization is radiometric calibration~\cite{lin2011revisiting, chakrabarti2014modeling}. Unlike our approach, radiometric calibration methods do not target a specific, well-defined color space, and do not address the problem of local processing.

\section{Our Framework}\label{sec:method}

This section describes our overall framework, including network architecture, dataset generation, and training details.

\subsection{Formulation} \label{subsec:formulation}

\newcommand{\im}{\mathbf{x}}
\newcommand{\imraw}{\im_{\mathrm{raw}}}

\newcommand{\imxyz}{\im_{\mathrm{xyz}}}
\newcommand{\imxyzgt}{\imxyz^*}
\newcommand{\imxyzpred}{\hat{\im}_{\mathrm{xyz}}}

\newcommand{\imglob}{\im_{\mathrm{glob}}}

\newcommand{\imsrgb}{\im_{\mathrm{srgb}}}
\newcommand{\imsrgbgt}{\imsrgb^*}
\newcommand{\imsrgbpred}{\hat{\im}_{\mathrm{srgb}}}

\newcommand{\imres}{\im_{\mathrm{res}}}

\newcommand{\pipe}{\mathcal{F}}
\newcommand{\invpipe}{\mathcal{G}}

\newcommand{\pipeglob}{\pipe_\mathrm{glob}}
\newcommand{\pipeloc}{\pipe_\mathrm{loc}}

\newcommand{\invpipeglob}{\invpipe_\mathrm{glob}}
\newcommand{\invpipeloc}{\invpipe_\mathrm{loc}}

\newcommand{\mat}{\mathbf{M}}
\newcommand{\matfwd}{\mat_\mathrm{fwd}}
\newcommand{\matinv}{\mat_\mathrm{inv}}

\newcommand{\norm}[1]{\left\lVert#1\right\rVert}
\newcommand{\abs}[1]{\left\lvert#1\right\rvert}

\newcommand{\kernelSixbyN}{\phi}
\newcommand{\reshape}{\psi}

Inside a camera imaging pipeline, a raw-RGB image $\imraw \in \mathbb{R}^{h \times w}$ undergoes a sequence of processing stages to be transformed to the final output sRGB image $\imsrgb \in \mathbb{R}^{h \times w \times 3}$, where $h$ and $w$ represent the image height and width, respectively.
%\begin{equation}
%\imsrgb = \pipe(\imraw) .
%\end{equation}
As mentioned earlier, the raw-RGB image $\imraw$ is in a camera-dependent color space that is linear with respect to scene light irradiance falling on the sensor.  One of the early steps in the camera processing pipeline is to convert the camera-dependent color space to a device-independent color space---namely, CIE XYZ.  Based on this observation, instead of modeling the whole pipeline back to the raw-RGB image, we choose to model an intermediate representation of the image in the CIE XYZ color space $\imxyz \in \mathbb{R}^{h \times w \times 3}$ that is still linear with respect to scene irradiance, but is in a canonical color space.  We are interested in the on-camera rendering procedures that map the CIE XYZ images into the final display-referred (i.e., photo-finished) sRGB color space. This operation can be described as
\begin{equation}
\imsrgb = \pipe(\imxyz).
\end{equation}

In our method, instead of relying on a single function to model the pipeline stages between sRGB and CIE XYZ, we decompose this mapping into two parts: 1) \textit{global processing}, denoted collectively as $\pipeglob(\cdot)$, that is globally applied to all image pixels and 2) \textit{local processing}, denoted collectively as $\pipeloc(\cdot)$, that represents local photo-finishing operations, such as local tone mapping and selective color adjustments. The forward pipeline from $\imxyz$ to $\imsrgb$ can be represented as a cascade of the global and the local processes.
% Forward model, global processing
The global processing stage is represented as
\begin{equation}
\matfwd = \pipeglob(\imxyz)  ,
\label{eq:matfwd}
\end{equation}
\begin{equation}
\imglob = \reshape \left( \matfwd \ \kernelSixbyN(\imxyz) \right) ,
\label{eq:imglob-fwd}
\end{equation}
%\begin{equation}
%\imsrgb = \pipeloc(\imglob) ,
%\label{eq:imsrgb-fwd}
%\end{equation}
where $\matfwd \in \mathbb{R}^{3 \times 6}$ is a global transformation matrix and $\imglob$ is the globally processed image layer. The operator $\kernelSixbyN(\cdot)$ reshapes the image to be $6 \times n$ where $n$ is the number of pixels in the image and each pixel is transformed from three to six dimensions: $[R, G, B] \rightarrow [R, G, B, R^2, G^2, B^2]$, while the operator $\reshape$ reshapes the image from $3 \times n$ back to $h \times w \times 3$.
We chose $\matfwd$ to be nonlinear to capture global color processing operations, such as gamma compression.

% Forward model, local processing
As most consumer cameras locally process the captured scene-referred images to improve the quality of final rendered images \cite{hasinoff2016burst}, such global color processing may not be able to effectively model the function $\pipe$. To that end, we use a residual learning mechanism where we model the residual layer $\imres$ between the locally and globally processed layers of the image as follows:

\begin{equation}
\imres = \pipeloc(\imglob) ,
\label{eq:imres-fwd}
\end{equation}%
\begin{equation}
\imsrgb = \imglob + \imres .
\label{eq:imsrgb-fwd}
\end{equation}%

% Inverse model
Now, the decomposition process applies the inverse process of Equations~\ref{eq:matfwd} -- \ref{eq:imsrgb-fwd} as follows:
\begin{equation}
\imres = \invpipeloc(\imsrgb) , \label{eq:imres-inv}
\end{equation}%
\begin{equation}
\imglob = \imsrgb - \imres , \label{eq:imglob-inv}
\end{equation}%
\begin{equation}
\matinv = \invpipeglob(\imglob), \label{eq:matinv}
\end{equation}%
\begin{equation}
\imxyz = \reshape \left( \matinv \ \kernelSixbyN(\imglob) \right) ,
\label{eq:imxyz-inv}
\end{equation}
where $\invpipeloc(\cdot)$ represents the inverse of residual local processing layer and $\invpipeglob(\cdot)$ is constrained to produce a global transformation matrix $\matinv \in \mathbb{R}^{3 \times 6}$ that represents the inverse global processing stage.

\begin{figure}[t]
\centering
\includegraphics[width=.7\linewidth]{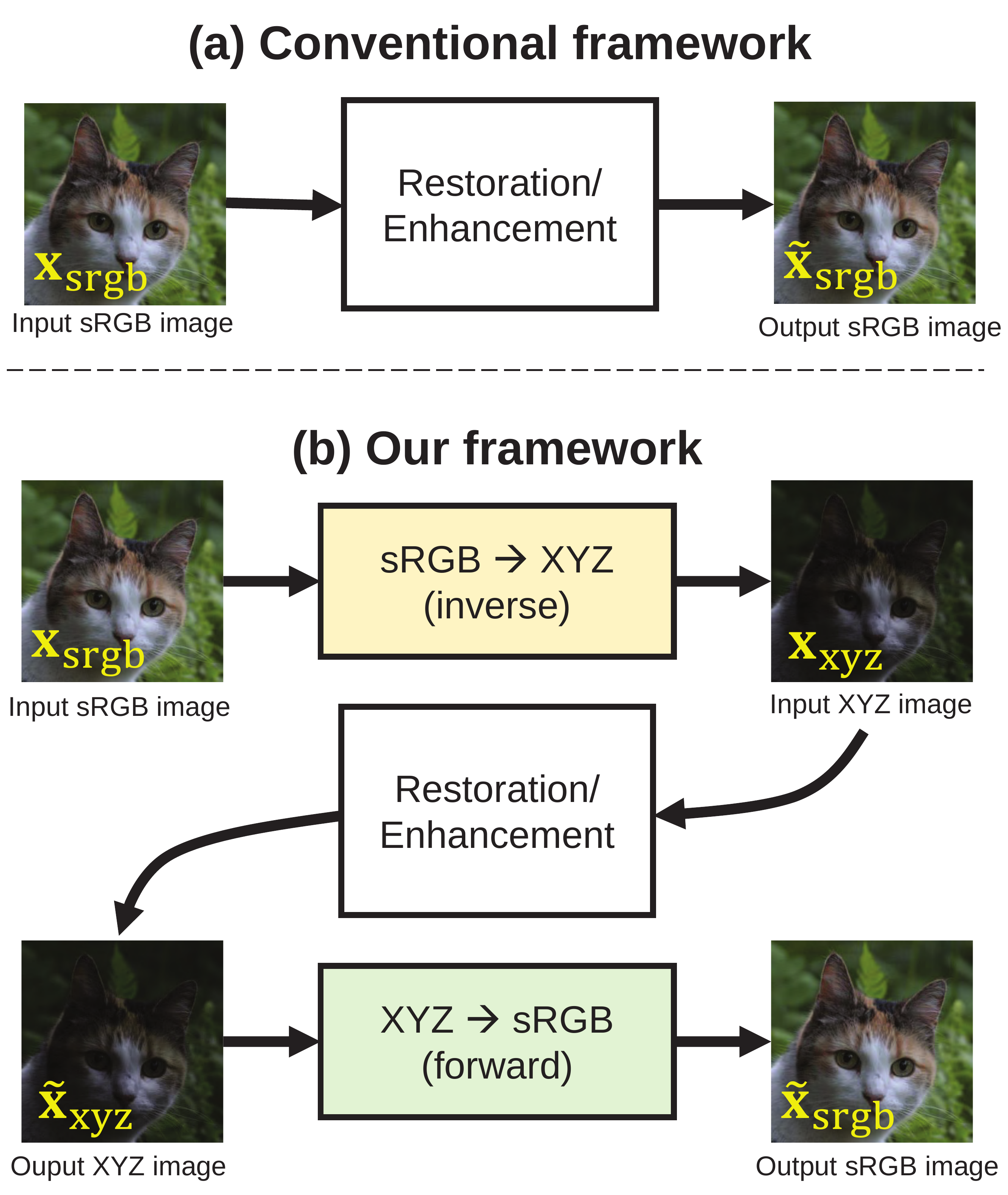}
\caption{An illustration of using our inverse and forward image processing pipelines in an sRGB image restoration/enhancement framework.}
\label{fig:framework}
\end{figure}

Our ultimate goal is to allow the manipulation of the reconstructed CIE XYZ image by arbitrary image restoration/enhancement algorithms between the inverse and forward pipeline stages (see Fig. \ref{fig:framework}).
It is, however, non-trivial to infer the inverse functions $\invpipeloc^{-1}(\cdot)$ and $\invpipeglob^{-1}(\cdot)$ to render back the reconstructed image, as its values may be changed by the image restoration or enhancement algorithms. To that end, we model each of $\pipeglob(\cdot)$, $\pipeloc(\cdot)$, $\invpipeglob(\cdot)$, and $\invpipeloc(\cdot)$ by a neural network.

\subsection{Network Design} \label{subsec:network}

\begin{figure*}[t]
\centering
\includegraphics[width=\linewidth]{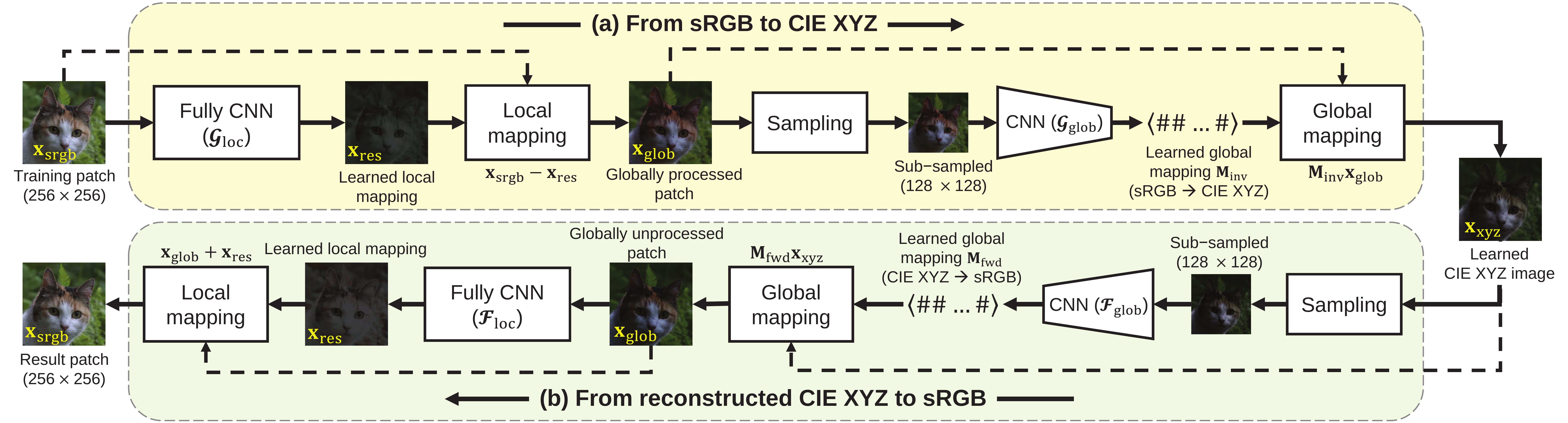}
\caption{Our CIE XYZ image pipeline. The upper part is the inverse pipeline that unprocesses an sRGB image into a CIE XYZ image. The lower part is the forward pipeline that processes a CIE XYZ image into its equivalent sRGB image. The full framework is trainable end-to-end. The CIE XYZ images are scaled 2x to aid visualization.}
\label{fig:overview}
\end{figure*}

% Network
Imitating this division of the camera imaging pipeline, we build our network architecture to include two sub-networks for modeling both the global and local processing parts for the forward and inverse directions of the imaging pipeline. As shown in Fig.~\ref{fig:overview}, we start with the inverse pipeline where the first part is a fully convolutional neural network (CNN) that models the local processing applied to an input non-linear image (i.e., sRGB image) by predicting the residual image $\imres$ (Equation~\ref{eq:imres-inv}). Once the local processing layer is predicted, it can be subtracted from the input image $\imsrgb$ to get the globally processed image $\imglob$ (Equation~\ref{eq:imglob-inv}). Then, $\imglob$ is fed to another sub-network that predicts a global transformation $\matinv$ that inverts $\imglob$ back to the linear CIE XYZ image $\imxyz$ (Equation~\ref{eq:imxyz-inv}). With this inverse pipeline, we decompose the input image $\imsrgb$ into two image layers, $\imres$ and $\imglob$, which represent local and  global processing, respectively, and finally output the linear CIE XYZ image $\imxyz$.

As discussed in Section~\ref{sec:intro}, there are computer vision tasks, such as image restoration, that are best processed in a linear image state.  A use case of the framework is to convert the input image $imxyz$, process the $imxyz$ image, and then render the image back.  In this scenario, after decomposing an image and applying an image restoration task to the linear XYZ image, we now need to merge these image layers back to produce the fully processed sRGB image. To model this forward pass of our pipeline, as shown in Fig.~\ref{fig:overview}, we use two sub-networks. The first sub-network predicts a global transformation $\matfwd$ that maps $\imxyz$ to $\imglob$ (Equation~\ref{eq:imglob-fwd}). The second sub-network predicts the residual local processing $\imres$ that needs to be applied to $\imglob$ to obtain the final sRGB image $\imsrgb$ (Equation~\ref{eq:imsrgb-fwd}). This framework is illustrated in Fig.~\ref{fig:framework} and compared to the conventional way of directly processing the sRGB image.

\begin{figure}[t]
\centering
\includegraphics[width=\linewidth]{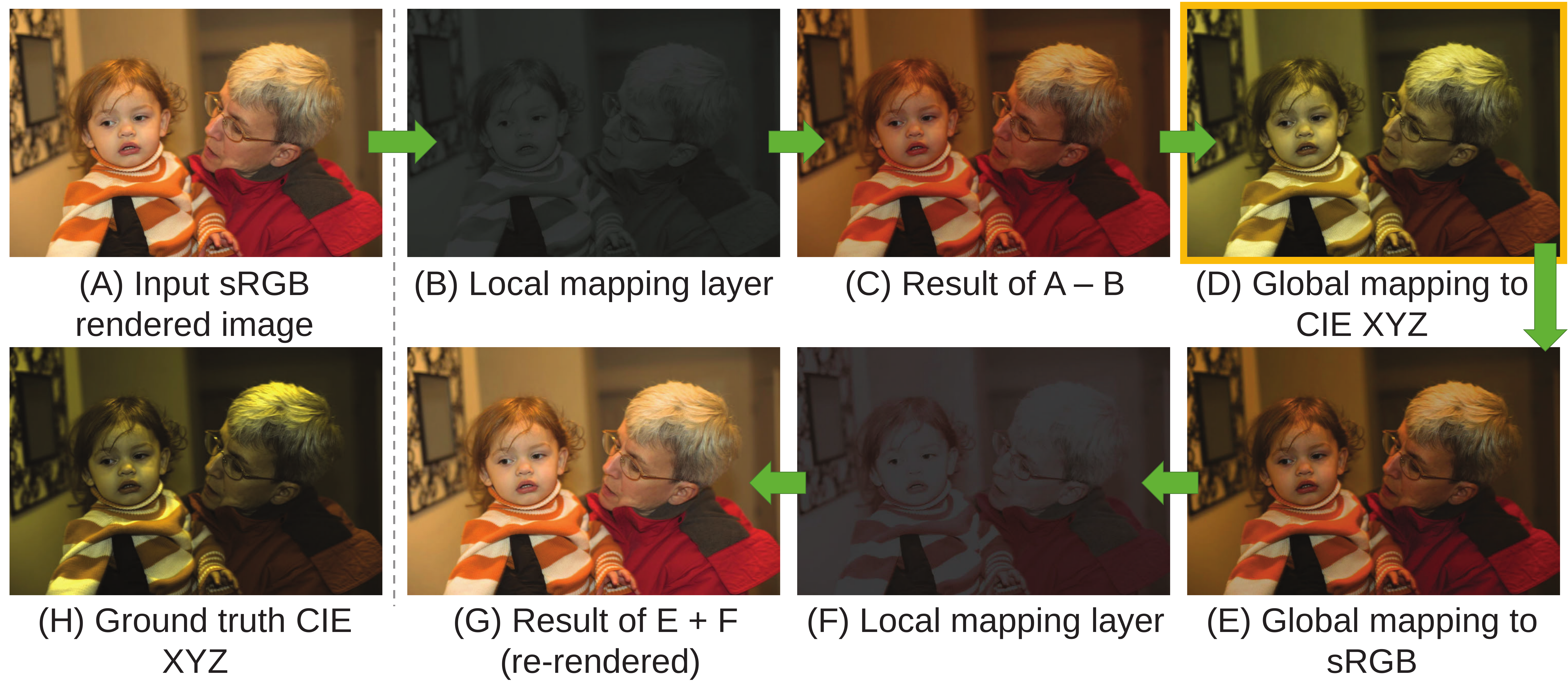}
\caption{Our inverse pipeline decomposites a given camera-rendered sRGB image into a local processed layer and the corresponding CIE XYZ image, while our forward pipeline maps the reconstructed CIE XYZ image to the sRGB color space in an inverse way of our decomposition. The shown image is taken from our testing set. To aid visualization, CIE XYZ images are scaled by a factor of two.}
\label{fig:analysis}
\end{figure}

% the joint learning of global and local layers
In order to allow the networks $\invpipeglob(\cdot)$ and $\invpipeloc(\cdot)$ to separate the globally and locally processed image layer without having ground truth for both $\imglob$ and $\imres$, we apply a scaling factor to the output of the local processing networks $\invpipeglob(\cdot)$, in both inverse and forward passes, such that the values of $\imres$ are much smaller than $\imglob$. In our experiments, we set this scaling factor to $0.25$. Fig. \ref{fig:analysis} shows an example of the output of each sub-network.

\subsection{Loss Function}
The objective of the whole network is to minimize the mean absolute error (MAE): 1) between the predicted XYZ image $\imxyzpred$ and its ground truth $\imxyzgt$ in the inverse pipeline and 2) between the predicted sRGB image $\imsrgbpred$ and its ground truth $\imsrgbgt$ in the forward pipeline, as follows:
\begin{equation}\label{eq:loss}
\lambda \abs{\imxyzpred - \imxyzgt} + \abs{\imsrgbpred - \imsrgbgt},
\end{equation}%
where $\lambda$ is a weighting factor that we use to deal with the fact that XYZ images generally have lower intensity compared to sRGB images; so this weight can balance the learning behavior between the forward and inverse pipelines. In our experiments, we set $\lambda = 1.5$.

\subsection{Sub-Networks Architecture} % move to supplemntal?
Our local processing sub-networks ($\pipeloc$ and $\invpipeloc$) each consist of 15 blocks of $3\!\times\!3$ convolutional (conv)--LReLU layers. Each conv layer has 32 output channels, with stride of 1 and padding of 1. The last layer of these sub-networks has a single conv layer with three output channels, followed by a $\tanh$ operator.
As our global processing sub-networks are not fully convolutional, we use a fixed size of input by introducing a differentiable subsampling module that uniformly subsamples $128\!\times\!128$ color values of the processed image by the previous sub-network.
Our global sub-network includes five blocks of $3\!\times\!3$ conv--LReLU--$2\!\times\!2$ max pooling layers. The conv layers have stride and padding of 1, while the max pooling layers have a stride factor of 2 with no padding. Then, we added a fully connected layer with 1024 output neurons, followed by a dropout layer with a factor of 0.5. The last layer of our global sub-network has a fully connected layer with 18 output neurons to formulate our $3\!\times\!6$ polynomial mapping function.
% parameters
Our entire framework is a light-weight model with a total of 2,697,578 learnable parameters ($\sim$11MB of memory) for both sRGB-to-XYZ and XYZ-to-sRGB models, and it is fully differentiable for end-to-end training.

\subsection{Dataset} \label{subsec:dataset}

To train our proposed model, we need a dataset of sRGB images with their corresponding linear images in the CIE XYZ color space. To do so, we start from raw-RGB images taken from the MIT-Adobe FiveK~\cite{fivek}. We then process the raw-RGB images twice to obtain both the sRGB and XYZ versions of each image. For processing raw-RGB images into the XYZ color space, we used the camera pipeline from~\cite{abdelhamed2018high}. This pipeline provides an access to the CIE XYZ values after processing the sensor raw-RGB using the color space transformation (CST) matrices provided with the raw-RGB image.
To obtain the camera-like sRGB images, we used the Adobe Camera RAW software development kit (SDK), which accurately emulates the nonlinearity applied by consumer cameras \cite{afifi2019color}. Our dataset includes $\sim$1,200 pairs of sRGB and camera CIE XYZ images. Our dataset will be publicly available upon acceptance.

\subsection{Training} \label{subsec:method-training}

We divided our dataset into a training set of 971 pairs, a validation set of 50 pairs, and a testing set of 244 pairs. We trained our framework in an end-to-end manner on patches of size $256\!\times\!256$ pixels randomly extracted from our training set, with a mini-batch of size 4. We applied random geometric augmentation (i.e., scaling and reflection) to the extracted patches.

Our framework was trained in an end-to-end manner for 300 epochs using Adam optimizer~\cite{kingma2014adam} with gradient decay factor $\beta_1= 0.9$ and squared gradient decay factor $\beta_2 =0.999$. We used a learning rate of $10^{-4}$ with a drop factor of 0.5 every 75 epochs. We added an $L_2$ regularization with a weight of $\lambda_{\mathrm{reg}} = 10^{-3}$ to our loss in Equation \ref{eq:loss} to avoid overfitting.

\section{Experimental Results}\label{sec:results}
In this section, we first validate the effectiveness of our proposed model in mapping from camera-rendered sRGB images to CIE XYZ, and processing CIE XYZ images back to sRGB. Next, we demonstrate our method's utility on a number of classical image restoration tasks, such as image denoising, motion deblurring, and image dehazing, that assume a linear relationship between the scene radiance and the recorded pixel intensity. Finally, we show that our linear space is also advantageous for a variety of image enhancement tasks.

\subsection{From Camera-Rendered sRGB to CIE XYZ, and Back} \label{subsec:results-XYZ}
We first verify our network's ability to unprocess sRGB images to CIE XYZ. We also demonstrate our ability to reconstruct from CIE XYZ back to sRGB. We test our mapping to sRGB using our reconstructed CIE XYZ results as a starting point, and also using the ground-truth CIE XYZ images. For comparison, we use the \textit{standard CIE XYZ mapping}~\cite{anderson1996proposal, ebner2007color}, which applies a simple 2.2 gamma tone curve, and the recent \textit{unprocessing technique (UPI)} from~\cite{Barron2019}. The UPI technique provides a proxy for the major procedures of the camera pipeline. For a fair comparison, we compare our results with results of UPI obtained at the CIE XYZ stage.

\begin{figure*}[t]
\centering
\includegraphics[width=0.999\linewidth]{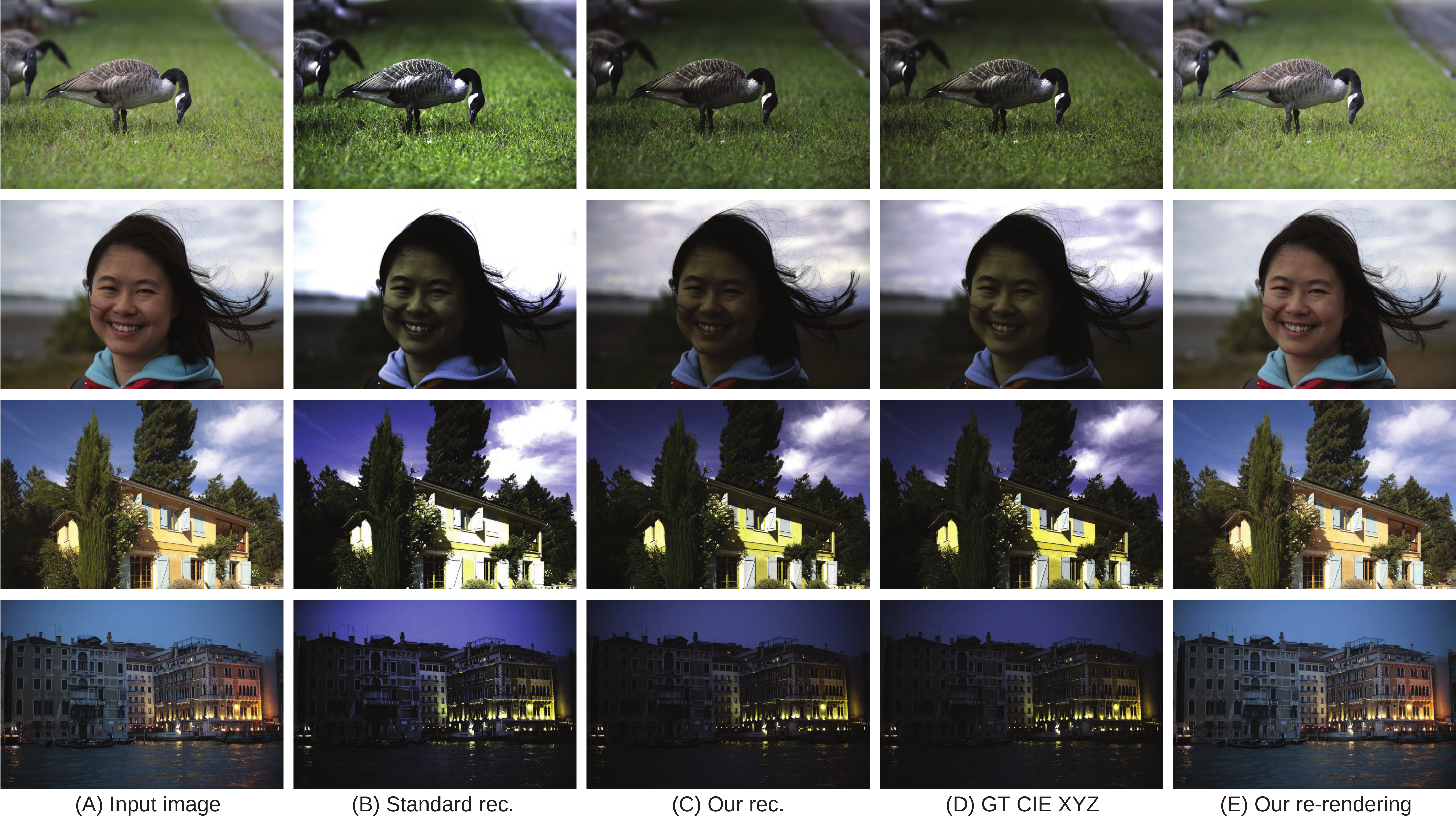}
\caption{Qualitative comparisons for CIE XYZ reconstruction and rendering. (A) The input sRGB rendered image. (B) Standard display-referred CIE XYZ reconstruction \cite{anderson1996proposal, ebner2007color}. (C) Our reconstruction. (D) The ground-truth scene-referred CIE XYZ image. (E) Our re-rendering result from the reconstructed CIE XYZ image. To aid visualization, CIE XYZ images are scaled by a factor of two. Input images are taken from the MIT-Adobe FiveK dataset \cite{fivek}.}
\label{fig:qualitative}
\end{figure*}

Table \ref{Table0} shows peak signal-to-noise ratio (PSNR) results averaged over 244 unseen testing images from the MIT-Adobe FiveK dataset~\cite{fivek}. The terms Q1, Q2, and Q3 refer to the first, second (median), and third quantile, respectively, of the PSNR values obtained by each method. For the standard XYZ, the results of mapping from the reconstructed CIE XYZ images back to sRGB are not reported because standard XYZ uses an invertible transform. The sRGB reconstruction error from the UPI model~\cite{Barron2019} is high due to the fact that the tone mapping is not perfectly invertible.
It can be observed from the results that we outperform both competing methods by a sound margin. Qualitative comparisons are provided in Fig. \ref{fig:qualitative}.

As shown in Table~\ref{Table0}, the mapping to sRGB from reconstructed CIE XYZ is better than mapping from ground-truth CIE XYZ. For our method, this behavior is expected because the forward model is trained on the reconstructed CIE XYZ, not the ground truth. Also, for the UPI method, as it is based on matrix inversion, the mapping from the reconstructed CIE XYZ makes the transformation more accurate than mapping from the ground truth.

% XYZ-sRGB mapping
\begin{table*}[t]
\centering
\caption{Results (in terms of PSNR) of camera-rendered sRGB $\leftrightarrow$ CIE XYZ mapping. We compare our results against the standard XYZ mapping (the 2.2 gamma tone curve)~\cite{anderson1996proposal, ebner2007color} and the recent unprocessing technique (UPI)~\cite{Barron2019}. Average PSNR (dB) results are reported on 244 unseen testing pairs (camera-rendered sRGB and corresponding CIE XYZ images) from the MIT-Adobe FiveK dataset~\cite{fivek}. We show results of mapping from both reconstructed (Rec.) CIE XYZ images and ground truth (GT) CIE XYZ images to the corresponding camera-rendered sRGB images. Highest PSNR values are shown in bold.\label{Table0}}
\scalebox{0.86}{
\begin{tabular}{|l|c|c|c|c|c|c|c|c|c|c|c|c|}
\hline
\multicolumn{1}{|c|}{} & \multicolumn{4}{c|}{\textbf{sRGB $\rightarrow$ XYZ}} & \multicolumn{4}{c|}{\textbf{Rec. XYZ $\rightarrow$ sRGB}} & \multicolumn{4}{c|}{\textbf{GT XYZ $\rightarrow$ sRGB}}\\ \cline{2-13}
\multicolumn{1}{|c|}{\multirow{-2}{*}{\textbf{Method}}} & \textbf{Avg.} & \textbf{Q1} & \textbf{Q2} & \textbf{Q3} & \textbf{Avg.} & \textbf{Q1} & \textbf{Q2} & \textbf{Q3} & \textbf{Avg.} & \textbf{Q1} & \textbf{Q2} & \textbf{Q3}  \\ \hline

Standard \cite{anderson1996proposal, ebner2007color} &  21.84 & 16.88 & 20.91 &  25.24 & - & - & -  & - & 22.22 & 19.19 & 21.79 &  24.37 \\ \hline

Unprocessing \cite{Barron2019}  & 22.19 & 19.31 & 22.12 & 24.75 & 37.72 & 37.78 & 40.56 & 41.88 & 18.04 & 15.67 & 17.79 & 20.02 \\ \hline

Ours & \textbf{29.66} & \textbf{23.77} & \textbf{29.57} & \textbf{34.71} & \textbf{43.82} & \textbf{41.43} & \textbf{43.94} & \textbf{46.58} & \textbf{27.44} & \textbf{23.57} & \textbf{28.32} & \textbf{30.88} \\ \hline
\end{tabular}
}

\end{table*}

We compare our proposed network against a U-Net-based baseline. This baseline consists of two U-Net-like \cite{unet} models trained in an end-to-end manner using the same training settings used to train our network (i.e., epochs, training patches, and loss function). Each U-Net model consists of a 3-level encoder/decoder with skip connections. The output channels of the first conv layer in the encoder unit has 28 channels. The two U-Net models have a total of 2,949,246 learnable parameters, compared to 2,697,578 learnable parameters in our network, and they were trained to map from sRGB to XYZ and from XYZ back to sRGB, similar to our model.

Table \ref{Table:unetcomparison} shows the results obtained by the U-Net baseline and our network on our testing set. 

\begin{table*}[]
\centering
\label{Table:unetcomparison}
\scalebox{1}{
\begin{tabular}{|l|c|c|c|c|c|c|c|c|}
\hline
\multicolumn{1}{|c|}{} & \multicolumn{4}{c|}{\textbf{sRGB $\rightarrow$ XYZ}} & \multicolumn{4}{c|}{\textbf{Rec. XYZ $\rightarrow$ sRGB}} \\ \cline{2-9}
\multicolumn{1}{|c|}{\multirow{-2}{*}{\textbf{Method}}} & \textbf{Avg.} & \textbf{Q1} & \textbf{Q2} & \textbf{Q3} & \textbf{Avg.} & \textbf{Q1} & \textbf{Q2} & \textbf{Q3}  \\ \hline

U-Net & 20.05 & 16.84 & 19.76 & 22.78 &  43.39 & 40.56 & 43.40 & 45.91 \\ \hline

Ours & \textbf{29.66} & \textbf{23.77} & \textbf{29.57} & \textbf{34.71} & \textbf{43.82} & \textbf{41.43} & \textbf{43.94} & \textbf{46.58}  \\ \hline

\end{tabular}
}

\caption{Comparison between our network and two U-Net models trained in an end-to-end manner to map from sRGB to CIE XYZ and back. Both networks, ours and the two U-Net models, have approximately the same number of learnable parameters and both were trained using the same training settings. The best PSNR (dB) values are shown in bold.}

\end{table*}

\subsection{Image Restoration Applications}  \label{subsec:results-applications}
As previously discussed, many image restoration tasks assume a linear image formation model and work best with linearized data. In the following sections, we apply our method to the problems of image denoising, motion deblurring, defocus blur detection, raw-RGB image reconstruction, and image dehazing. We show improvement in performance from working on CIE XYZ images compared to working directly with sRGB images or applying existing linearization methods. We would like to highlight that our objective is \emph{not} to outperform the state-of-the-art. Instead, our goal is to show that having selected a particular algorithm for a given task, its performance improves when applied to our linearized images as compared to using sRGB images, or employing existing linearization approaches.

\subsubsection{Image Denoising}

We first apply our decomposition method to the task of image denoising. Image denoising algorithms can perform better on linear images because of the simplicity of the linear signal-dependent noise model~\cite{plotz2017benchmarking,abdelhamed2018high}.  We demonstrate this behavior using our CIE XYZ images for two denoising methods:  BM3D~\cite{dabov2007image} and DnCNN~\cite{zhang2017beyond}. The first is statistics-based while the latter is learning-based. Both methods are well established in the image denoising literature. We evaluated the chosen methods on the SIDD-Validation dataset~\cite{abdelhamed2018high}.

We compare denoising CIE XYZ images from our method against the standard XYZ linearization~\cite{anderson1996proposal, ebner2007color}. First, the sRGB images are linearized, denoising is performed, and the denoised images are mapped back to sRGB. As a baseline, we also apply denoising directly in the sRGB space. For the BM3D denoiser, we used its color-variant version---namely, CBM3D~\cite{dabov2007image}. For DnCNN, we used the pre-trained model from~\cite{zhang2017beyond}, denoted as DnCNN-P. We also re-trained DnCNN models on images processed with both linearization methods, denoted as DnCNN-R.

The evaluation is based on the PSNR (dB) and structural similarity index (SSIM)~\cite{wang2004image}, both in the sRGB color space. The results are shown in Table~\ref{tab:denoising}. Both CBM3D and DnCNN-P show significant denoising improvement when applied to images unprocessed by our method into the CIE XYZ space compared to linearized images obtained by the standard XYZ linearization. DnCNN-R using our CIE XYZ images gains more improvement in terms of SSIM, but is still comparable to using standard XYZ images in terms of PSNR. Also, we can see that denoising in our reconstructed CIE XYZ space is better compared to denoising directly in the sRGB color space.

% denoising
\begin{table*}[t]
\centering
\scalebox{0.83}{
\begin{tabular}{|l|c|c|c|c|c|c|}
\hline

\multicolumn{1}{|c|}{} & \multicolumn{2}{c|}{\textbf{CBM3D}} & \multicolumn{2}{c|}{\textbf{DnCNN-P}} & \multicolumn{2}{c|}{\textbf{DnCNN-R}} \\ \cline{2-7}

\multicolumn{1}{|c|}{\multirow{-2}{*}{\textbf{Method}}} & \textbf{PSNR} & \textbf{SSIM} & \textbf{PSNR} & \textbf{SSIM} & \textbf{PSNR} & \textbf{SSIM}  \\ \hline

XYZ - Standard & 33.84 & 0.8733 &  26.23  &  0.5936  & \textbf{33.92} &  0.7833  \\ \hline

XYZ - Ours & \textbf{35.49} & \textbf{0.9245} & \textbf{27.71} & \textbf{0.6623} & 33.00 & \textbf{0.8865} \\ \hline

sRGB & 34.44 & 0.9021 & 25.31 & 0.5578 & 30.72 & 0.7943\\ \hline

\end{tabular}
}
\caption{Results of image denoising performed on the SIDD-Validation set~\cite{abdelhamed2018high} after unprocessing the images using our method compared to standard linearization~\cite{anderson1996proposal, ebner2007color}. The last row is the result of denoising directly in the sRGB space. All PSNR and SSIM values are calculated in the sRGB space. CBM3D and DnCNN-P show significant improvement in denoising using images unprocessed by our method.\label{tab:denoising}}
\end{table*}

\subsubsection{Motion Deblurring}

Traditionally, a motion blurred image $G$ is represented using a linear image formation model as the convolution between an underlying sharp image $F$, and a motion blur kernel $H$~\cite{Fergus06removingcamera}---namely, $G = F \otimes H$, where $\otimes$ denotes the convolution operation. The blur kernel $H$ is the same across the image.

However, this linear relationship between $G$ and $F$ does not hold for blurred photo-finished sRGB images. Tai et al.~\cite{tai} showed that the blur kernel $H$ varies spatially across the image due to the non-linear stages of the camera pipeline, making motion deblurring significantly more challenging. Therefore, it is advantageous to deblur the image in a linear scene-referred space.

\begin{figure}[t]
\includegraphics[width=0.98\linewidth]{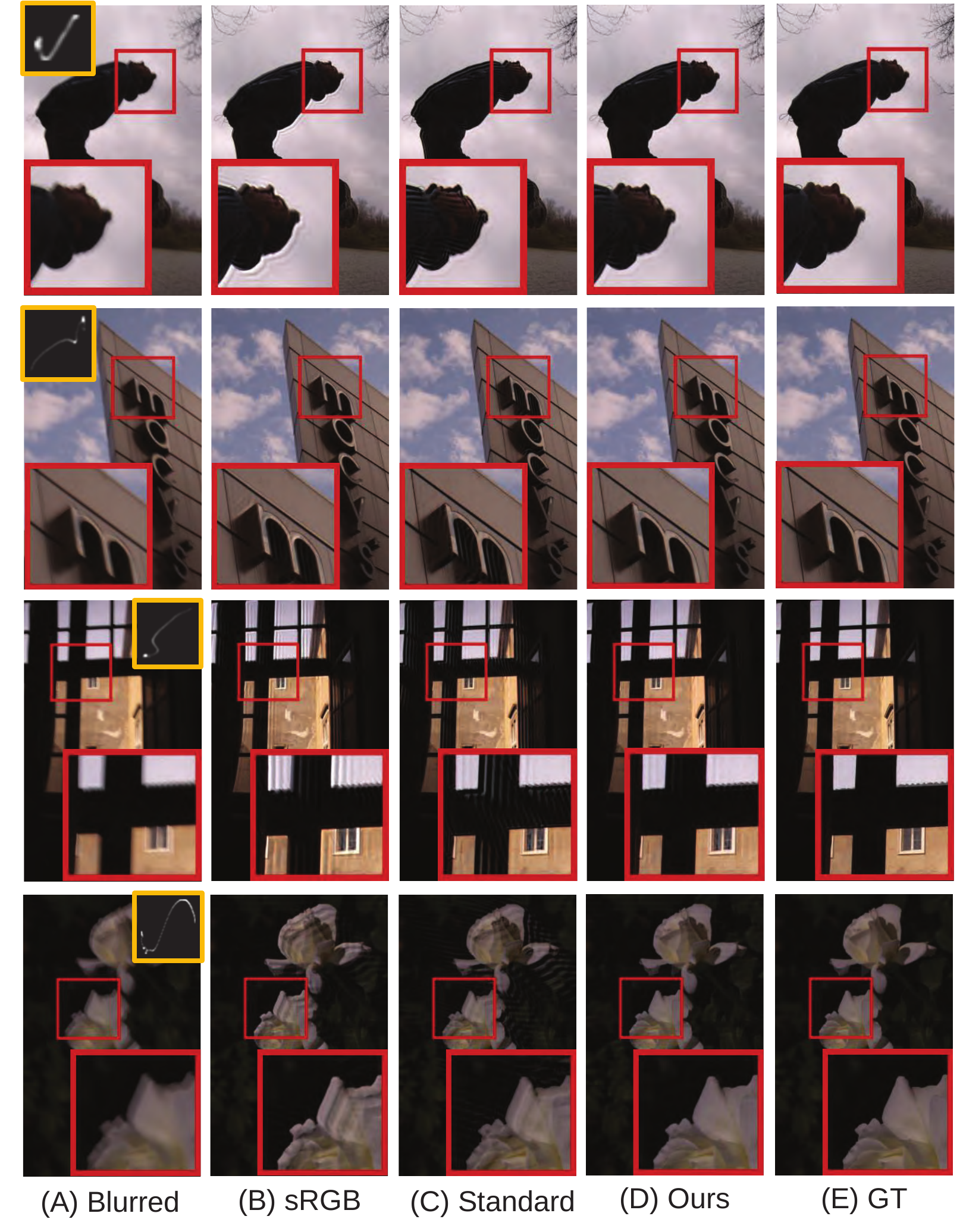}
\caption{Qualitative results for motion deblurring application.
(A) The blurred input image and the corresponding motion blur kernel. (B-D) Deblurring results in sRGB, standard XYZ, and our proposed CIE XYZ color space, respectively. The shown deblurring results are obtained by the non-blind image deblurring algorithm of Krishnan and Fergus~\cite{nips09}. (E) The ground-truth sharp image. Images are taken from the MIT-Adobe FiveK dataset~\cite{fivek}, and the motion blur kernels are the four kernels used in the benchmark motion blur dataset of Lai et al.~\cite{MBdataset}.
\label{fig:motion_blur}}
\end{figure}

We select the statistics-based non-blind motion deblurring method of Krishnan and Fergus~\cite{nips09}, and evaluate its performance in our linearized color space, as well as other spaces, such as nonlinear sRGB and standard XYZ~\cite{anderson1996proposal, ebner2007color}. For this experiment, we chose the same 244 unseen testing images from the MIT-Adobe FiveK dataset~\cite{fivek} used earlier in Section \ref{subsec:results-XYZ}. To generate motion blurred sRGB images, we blur each ground-truth raw-RGB image with a random kernel selected from one of the four kernels in the widely used motion blur benchmark dataset of Lai et al.~\cite{MBdataset}. The blurred raw-RGB images are then rendered to sRGB using the camera pipeline simulator of~\cite{abdelhamed2018high}. Fig. \ref{fig:motion_blur} shows four representative deblurring results, one for each motion kernel from the dataset of~\cite{MBdataset}. It can be clearly observed from the zoomed-in regions that deblurring in the sRGB and the standard CIE XYZ color spaces produces a lot of ringing artifacts, particularly for larger blur kernels (e.g., the examples in the last two rows). In comparison, our results are sharper and less prone to ringing. Deblurring in our linear space produced an average PSNR of 30.26 dB, whereas sRGB and standard XYZ yielded only 29.41 dB and 28.68 dB, respectively---a clear demonstration of the utility of our proposed framework for motion deblurring.

\subsubsection{Defocus Blur Detection}\label{secdefocus}

\begin{figure}
\centering
\includegraphics[width=0.95\linewidth]{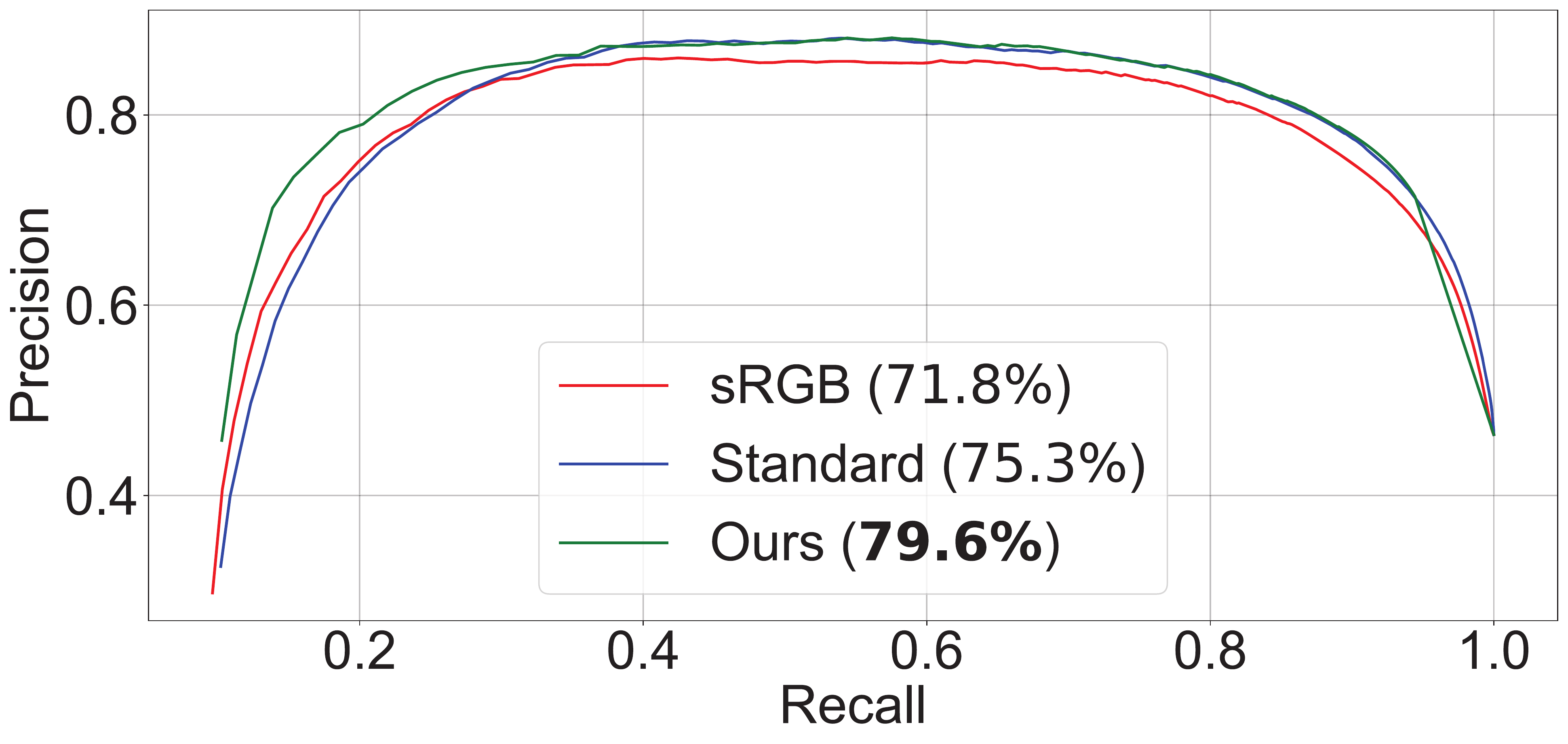}
\caption{The precision-recall (PR) comparison of training light UNet on data from three different color spaces: sRGB, standard linearized CIE XYZ~\cite{anderson1996proposal, ebner2007color}, and our linearized CIE XYZ. The average accuracy for each model is shown in the plot's legend. Our linear space achieves the best PR curve and the highest accuracy.}
\label{fig:defocusRes}
\end{figure}

\begin{figure}
\centering
\includegraphics[width=\linewidth]{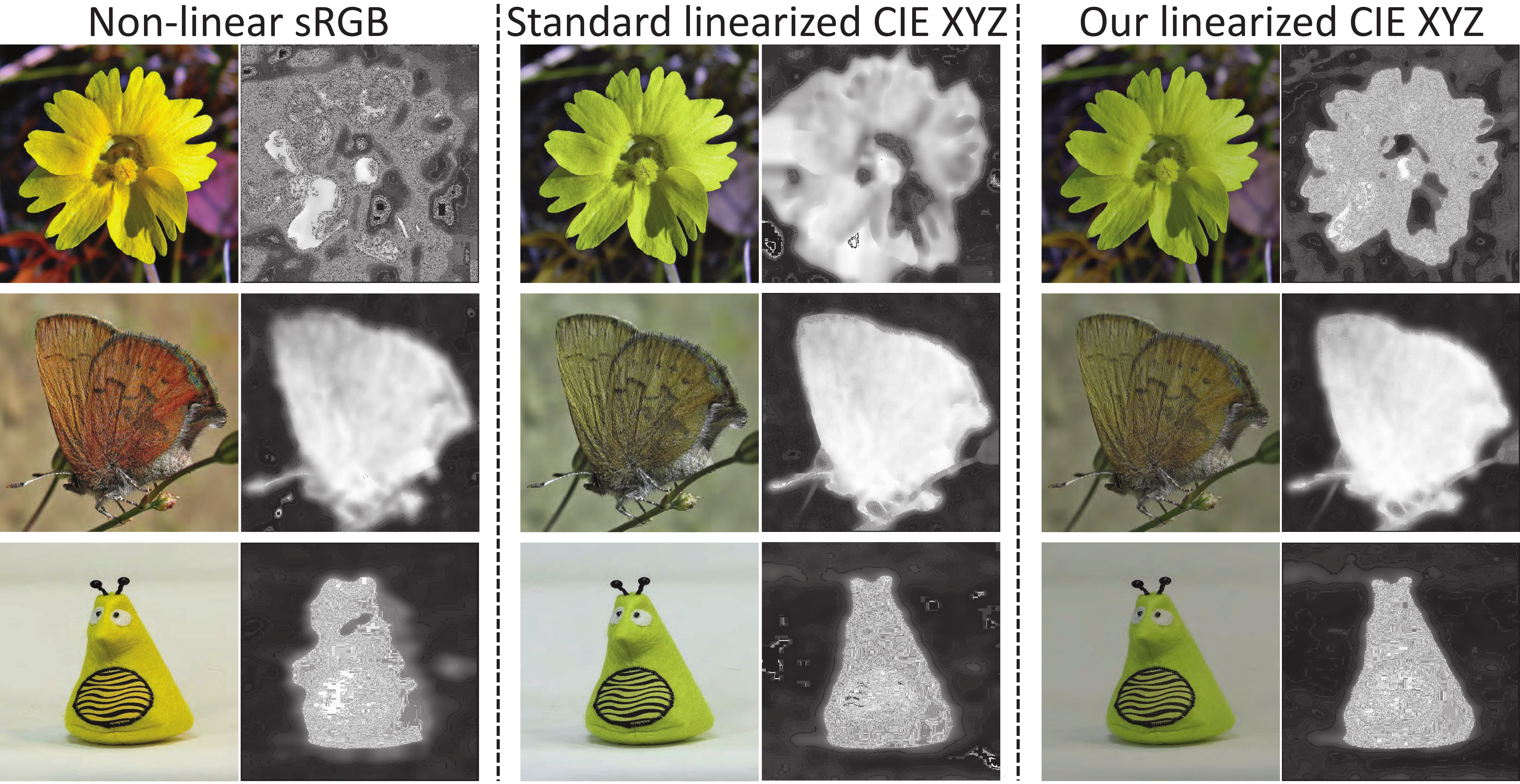}
\caption{Qualitative results of three light UNets trained on three different color spaces for the task of defocus map estimation. Training the light UNet using our linearized CIE XYZ images gives better visual results as shown in the third column. The CIE XYZ input images are gamma corrected for better visualization.}
\label{fig:defocusQualRes}
\end{figure}

Defocus detection is the problem of detecting the image pixels that are out of focus. This problem is important to many computer vision tasks, such as non-blind defocus deblurring, image refocusing, depth estimation, and 3D reconstruction. The methods targeting this problem are well established~\cite{golestaneh2017spatially, shi2014discriminative, tang2019defusionnet, yi2016lbp, zhao2018defocus, zhao2019enhancing}. In particular, defocus detection methods output a binary mask of a given input image such that the out-of-focus pixels are zeros and the rest are ones.

We examine the advantage of detecting out-of-focus pixels in our linearized CIE XYZ color space compared to other spaces---namely, non-linear sRGB and standard linearized CIE XYZ. To this aim, we introduce a light UNet-inspired architecture. We train three models using image patches from three different color spaces following the same training procedure. We use the well-known blur detection dataset~\cite{shi2014discriminative} to train and test the models. In the plot of Fig.~\ref{fig:defocusRes}, we present the precision-recall comparison along with the pixel binary classification accuracy. In addition to the quantitative results, qualitative results are presented in Fig.~\ref{fig:defocusQualRes}. The results in Fig.~\ref{fig:defocusRes} and Fig.~\ref{fig:defocusQualRes} demonstrate that our linearization is a better color space to perform defocus blur detection compared to other color spaces. 

\begin{figure}[t]
\centering
\includegraphics[width=\linewidth]{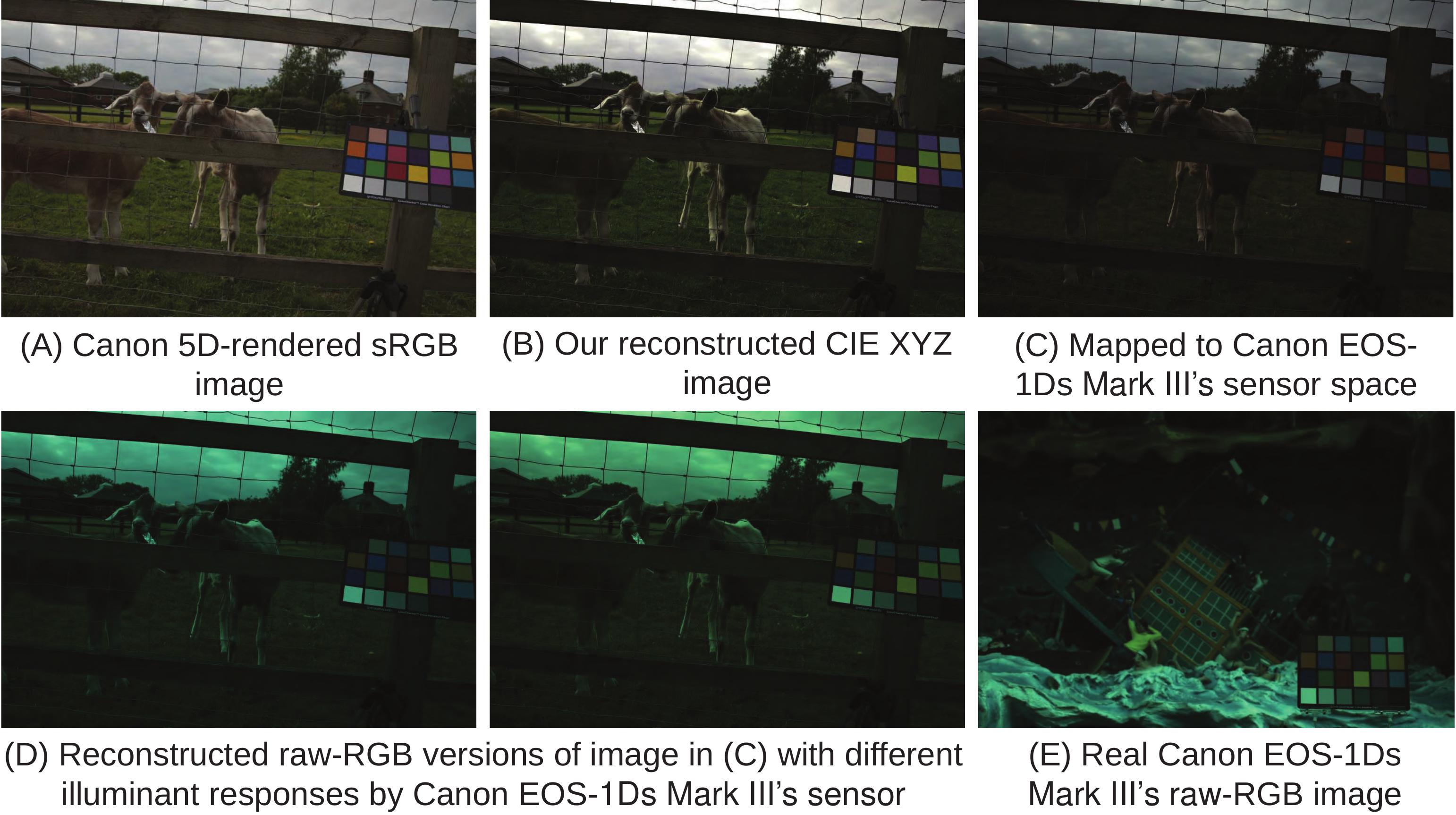}
\caption{Sensor raw-RGB image reconstruction. (A) An sRGB image rendered by Canon 5D from Gehler-Shi \cite{gehler2008bayesian}. (B) Our reconstructed CIE XYZ image. (C) Our reconstructed raw image in the raw-RGB space of the Canon EOS-1Ds Mark III. (D) Two generated raw-RGB images with different illuminant responses in the Canon EOS-1Ds Mark III's sensor space. (E) A real raw-RGB image captured by the Canon EOS-1Ds Mark III taken from the eight-camera NUS dataset \cite{cheng2014illuminant}. %
%Abdo: why we need to show E?
To aid visualization, the shown images are scaled by a factor of two.}
\label{fig:raw_reconstruction}
\end{figure}

\subsubsection{Raw-RGB Image Reconstruction}\label{secrawrec}
One of the advantages of accurately reconstructing scene-referred images is the ability to map the reconstructed images further into a sensor raw-RGB space. Specifically, we can synthetically generate raw-RGB images in any target sensor space by capturing an image with a color rendition calibration chart placed in the scene. The captured image is saved in both the camera's sensor raw-RGB space and the camera-rendered sRGB color space. As the CIE XYZ space is defined for correctly white-balanced colors, we first correct the white balance of the raw-RGB image using the color rendition chart. We then reconstruct the XYZ image using our XYZ network and compute a $3\!\times\!3$ matrix to map our reconstructed image into the sensor space. We refer to this matrix as the XYZ$\rightarrow$raw matrix. %
%Abdo: this XYZ->raw matrix will be different under different illuminations, right?
This calibration matrix is then used to map any arbitrary image into this sensor space by first reconstructing the corresponding XYZ image, followed by mapping it into the sensor space. The assumption here is that as our method achieves superior linearization to the available solutions (see Table \ref{Table0}), this calibration process would result in a better sRGB$\rightarrow$raw-RGB mapping.

To validate this assumption, we compare between the raw-RGB reconstruction based on our reconstruction against the raw-RGB reconstruction that is computed based on the standard XYZ mapping method~\cite{anderson1996proposal, ebner2007color}. We examine the data augmentation task for the illuminant estimation problem. Scene illuminant estimation is a well-studied problem in computer vision literature. Briefly, we can describe the illuminant estimation problem as follows. Given a linear raw-RGB image $\mathbf{I}_{\text{raw}}$ captured by a specific camera sensor, the goal is to determine a 3D vector $\pmb{\ell}$ that represents the illuminant color in the captured scene. Recent work achieves promising results using deep learning to estimate the illuminant vector $\pmb{\ell}$ by training deep models that can be later used in the inference phase to estimate illumination colors of given testing images captured by the same sensor used in the training stage \cite{gehler2008bayesian}.

\begin{figure}[t]
\centering
\includegraphics[width=\linewidth]{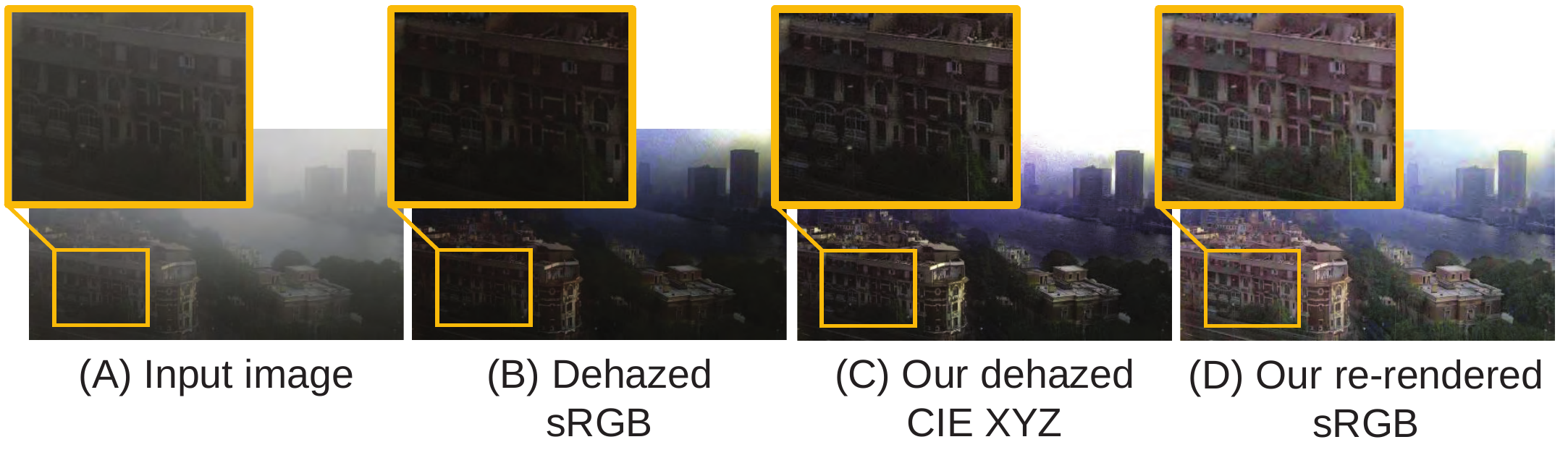}
\caption{Dehazing is one of the potential applications that can benefit from our image unprocessing method. (A) The input image taken from Flickr (by Mike Rivera, CC BY-NC-SA 2.0). (B) Dehazing applied in the sRGB space. %
%(C) Our  reconstructed CIE XYZ image of the sRGB image in (A).
(C) Dehazing applied to our CIE XYZ image. (D) Our final result in the sRGB space. In this example, we used the dehazing method from \cite{he2010single}.}
\label{fig:dehaze}
\end{figure}

\begin{figure}
\centering
\includegraphics[width=\linewidth]{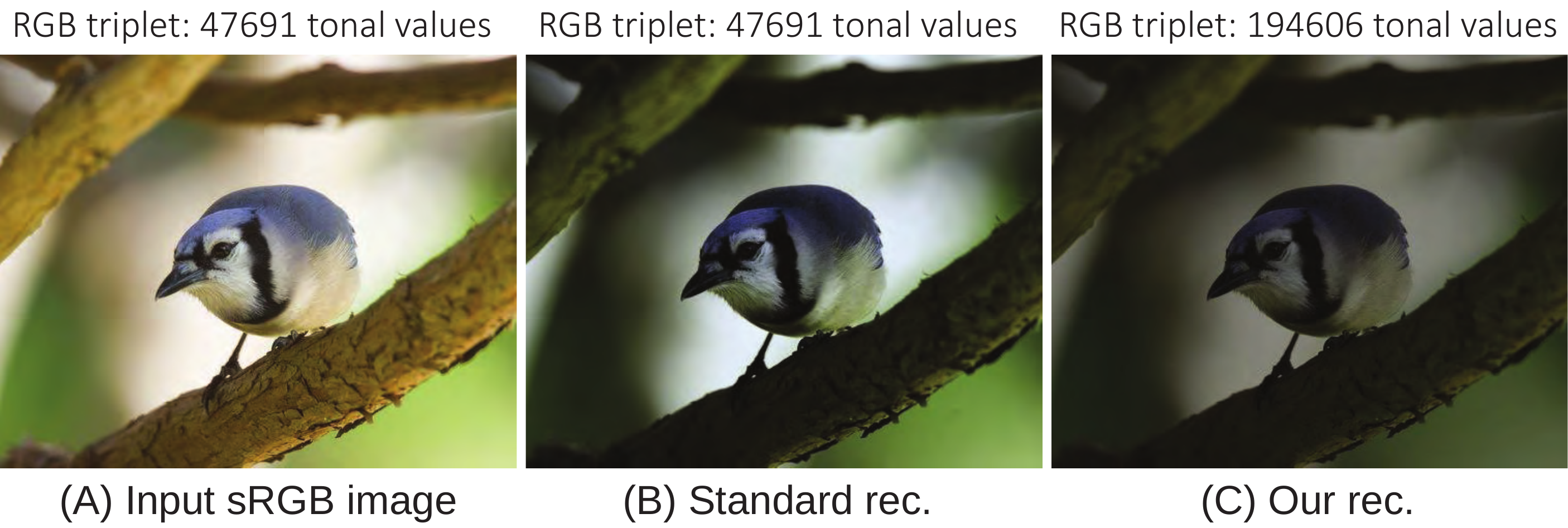}
\caption{Our XYZ reconstruction provides a wider range of tonal values compared to the standard CIE XYZ mapping~\cite{anderson1996proposal, ebner2007color}. (A) The input sRGB image. (B) Standard XYZ reconstruction \cite{anderson1996proposal, ebner2007color}. (C) Our XYZ reconstruction. The input image is taken from~\cite{WahCUB_200_2011}.}
\label{fig:tonal_values}
\end{figure}

\begin{figure*}[t]
\centering
\includegraphics[width=\linewidth]{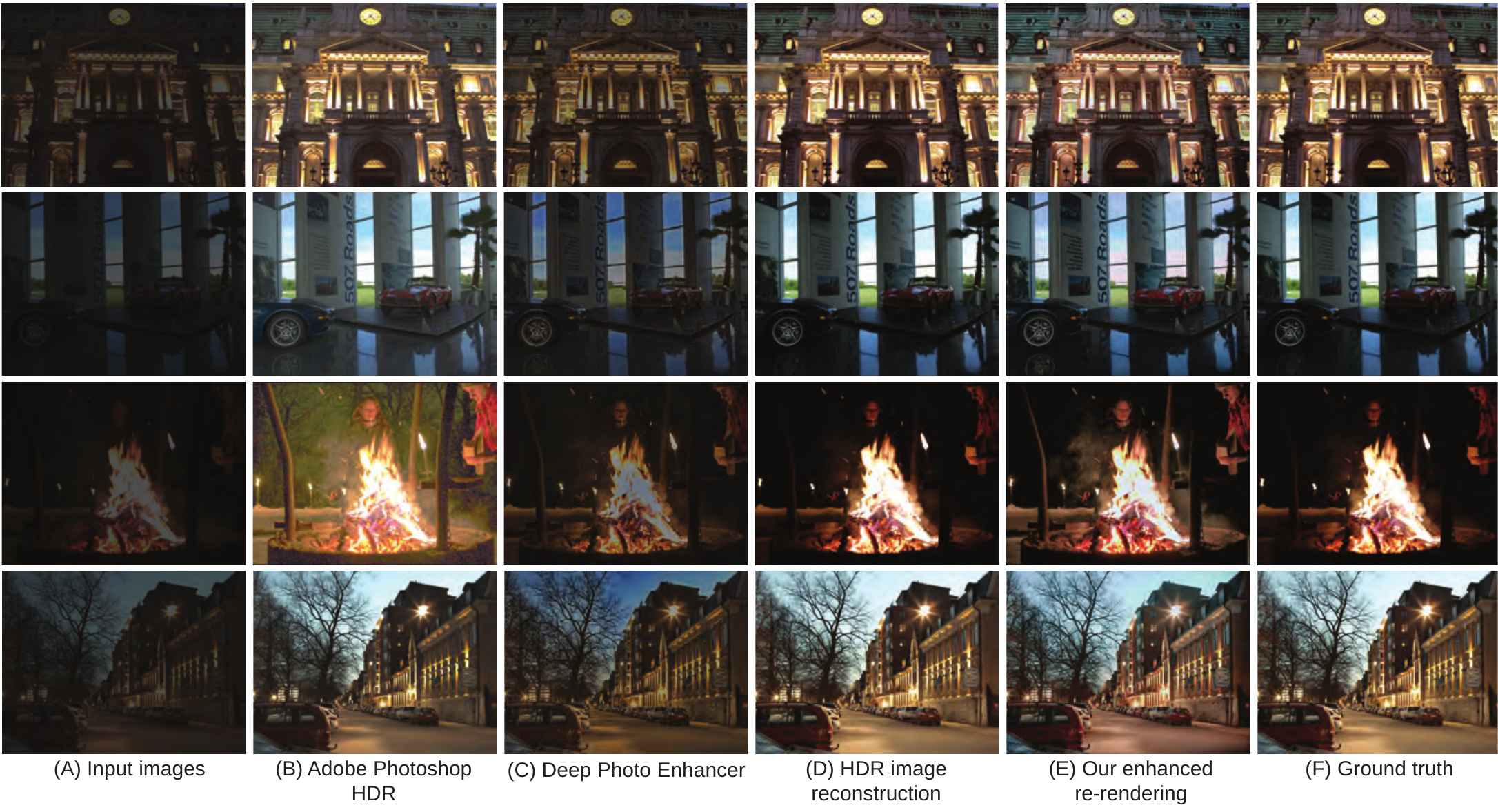}
\caption{(A) The input sRGB rendered image. (B) Adobe Photoshop HDR results. (C) Deep Photo Enhancer results \cite{chen2018deep}. (D) HDR result of ~\cite{eilertsen2017hdr}. (E) Our re-rendered images after photo-finishing enhancement. (F) Ground-truth images. Input images are taken from~\cite{eilertsen2017hdr}.}
\label{fig:HDR}
\end{figure*}

There is currently a challenge in the available datasets for the illuminant estimation task, which is the limited number of available training images captured by the same sensor---for example, the eight-camera NUS dataset \cite{cheng2014illuminant}, one of the common datasets used for illuminant estimation, has 200 images on average for each camera sensor. In this experiment, we examine our raw-like reconstructed images to serve as a data augmenter to train deep learning models for illuminant estimation. Specifically, we train a simple deep learning model to estimate the scene illuminant of a given raw-RGB image captured by Canon EOS-1Ds Mark III \cite{cheng2014illuminant}. There are only 256 original raw-RGB images captured by Canon EOS-1Ds Mark III in the NUS dataset \cite{cheng2014illuminant}. For each image, there is a ground-truth scene illuminant vector extracted from the color rendition chart. During training and testing processes, the color chart is masked out in each image to avoid any bias. To augment the data, we first computed the $3\!\times\!3$ XYZ$\rightarrow$raw calibration matrix as described earlier for our XYZ reconstruction and the standard XYZ mapping. This reconstruction process was performed using a single image captured by the Canon EOS-1Ds Mark III camera with a color rendition chart. %
% Abdo: so we use one image to compute the XYZ->RAW matrix?
Afterwards, we used 3,752 white-balanced camera-rendered sRGB images captured by ten different camera models other than our Canon EOS-1Ds Mark III. These images were taken from the Rendered WB dataset \cite{afifi2019color}. Each sRGB image is converted to the CIE XYZ space using our method and the standard XYZ mapping, followed by mapping each reconstructed image to the Canon EOS-1Ds Mark III sensor space using the calibration matrix computed for each XYZ reconstruction method, respectively.

As the calibration matrices map from the reconstructed XYZ space to the white-balanced sensor raw-RGB space, we can apply illuminant color casts, randomly selected from the ground-truth illuminant vectors provided in the Canon EOS-1Ds Mark III's set, to synthetically generate additional training data to train the deep model. Fig. \ref{fig:raw_reconstruction} shows an example.  This process is inspired by previous work in \cite{lou2015color, fourure2016mixed}, which randomly selected illuminant vectors from the ground-truth set and applied chromatic adaptation to augment the training set. These methods, however, use the same images (256 images in the case of the Canon EOS-1Ds Mark III's set) without introducing new image content to the trained model.

We randomly selected 50 testing images from the original 256 images provided in the NUS dataset for the Canon EOS-1Ds Mark III camera. We fixed this testing set over all experiments and excluded these images from any training processes. Table \ref{table:illuminant_estimation} shows the angular error of the trained model using the following training sets: (i) real training data, (ii) reconstructed raw-like images using the standard XYZ mapping, (iii) real training data and reconstructed raw-like images using the standard XYZ mapping, (iv) reconstructed raw-like images using our XYZ reconstruction, and (v) real training data and reconstructed raw-like images using our XYZ reconstruction. As can be seen, the best results were obtained by using our raw-like reconstruction and real training data. Notice that training only on our raw-like reconstruction gives better results compared with the results obtained by training on real data or reconstructed raw-like images using the standard XYZ mapping.

\begin{figure}[]
\includegraphics[width=\linewidth]{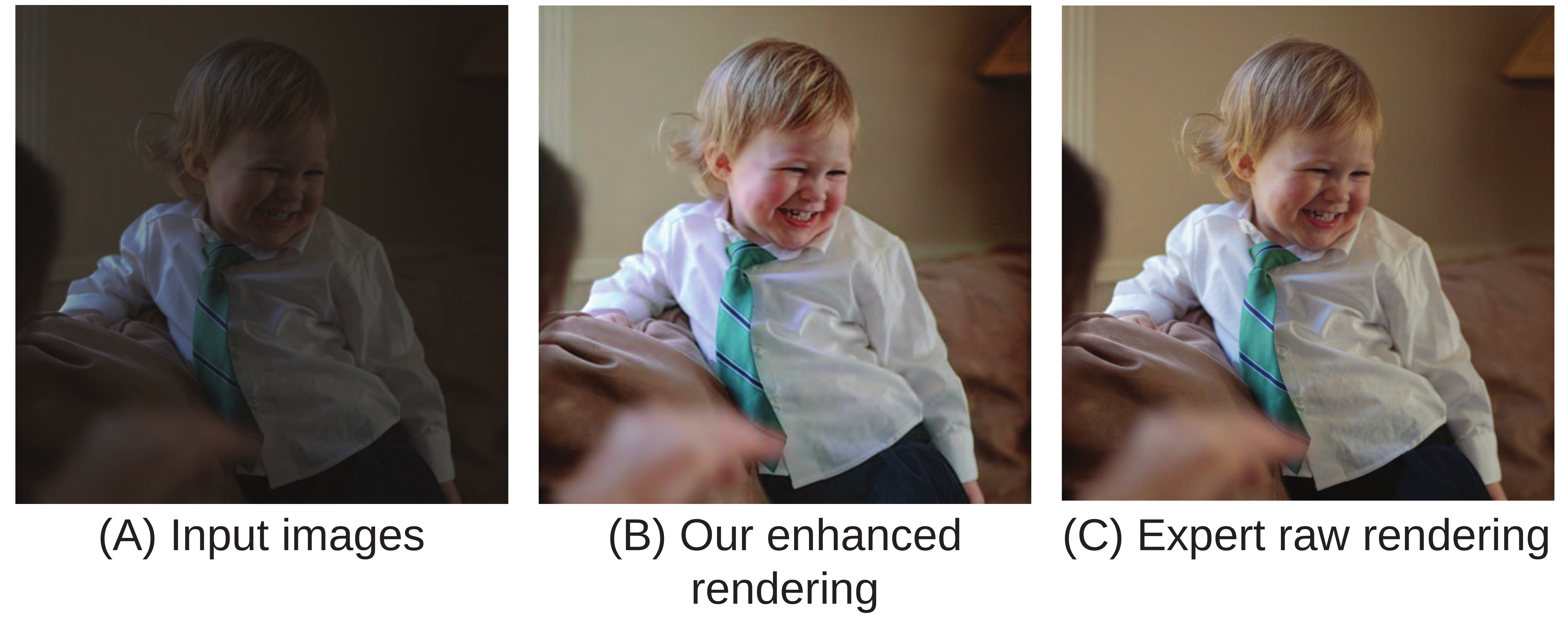}

  \caption{Example from the under-exposure testing set \cite{wang2019underexposed}. (A) The input image. (B) Our enhanced rendered image. (C) The expert-retouched image. \label{fig:low-light}}%
\end{figure}

\subsubsection{Dehazing}
% no barron, no standard XYZ, only sRGB
% and only one image
A hazy image is expressed using a linear model as  $\mathbf{I}(\mathbf{x})=\mathbf{J}(\mathbf{x})t(\mathbf{x})+\mathbf{A}(1-t(\mathbf{x}))$~\cite{he2010single}, where $\mathbf{I}$ is the observed intensity, $\mathbf{J}$ is the scene radiance, $\mathbf{A}$
is the global atmospheric light, and $t$ is the medium transmission describing the portion of the light that is not scattered and reaches the camera. Just as with motion deblurring, this linear relationship is broken by the camera's photo-finishing stages. Therefore, it is desirable to perform dehazing on linearized images.
In Fig. \ref{fig:dehaze}, we show the result of dehazing an sRGB image versus dehazing our linear CIE XYZ image and then re-rendering to sRGB. The improvement in visual quality can be clearly observed from the zoomed-in regions.

\subsection{Photo-Finishing Applications}
Many photographers prefer to edit photographs in the linear raw-RGB sensor space rather than the nonlinear 8-bit sRGB space, due to the fact that raw-RGB images provide higher tonal values compared to sRGB camera-rendered images \cite{schewe2015digital}. Similar to the raw-RGB space, the CIE XYZ space is linear scene-referred with higher tonal values compared to the final sRGB space. Thus, we can also benefit from our linear CIE XYZ space for image enhancement tasks.

\subsubsection{Low-Light Image Enhancement}
In this set of experiments, we present a set of simple operations that can achieve results on par with recent methods designed for low-light image enhancement. Specifically,  we apply the following set of heuristic operations to perform low-light image enhancement. As our reconstructed XYZ image has a wider range of tonal values (see Fig. \ref{fig:tonal_values}), we apply a set of synthetic digital gains to simulate multi-exposure settings. This simulation does not introduce any new information that did not exist in the original image; however, it allows us to better explore the range of tonal values provided in our reconstructed image---we can think of this operation as an ISO gain that is applied on board cameras to amplify the captured image signal. To that end, we multiply the reconstructed image by four different factors. These factors can be tuned in an interactive manner based on each image, but we preferred to fix these hyperparameters over all experiments. In particular, we multiplied our reconstructed XYZ image by (0.1, 1.4, 2.7, 4.0) to generate four different versions of our reconstructed XYZ image. Following this, we apply an off-the-shelf exposure-fusion algorithm~\cite{mertens2007exposure} to create the modified XYZ layer. To enhance the local details, we apply a local details enhancement method~\cite{paris2011local} on our forward local sRGB reconstructed layer. Fig. \ref{fig:HDR} shows examples of our results. As can be seen, we achieve on par results with state-of-the-art methods designed specifically for the given image enhancement task.

\begin{figure*}[]
\centering
\includegraphics[width=\linewidth]{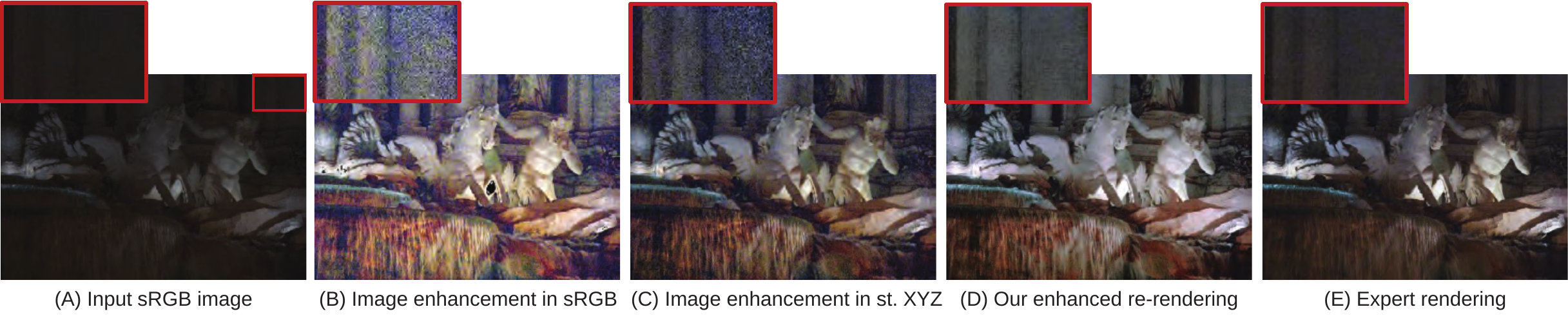}
\caption{(A) The input image. (B) Image enhancement in sRGB. (C) Image enhancement in standard XYZ reconstruction. (D) Our enhanced re-rendering. (E) Expert enhancement. The enhancement is based on fusion of ``multi-exposed'' images \cite{mertens2007exposure} and local details enhancement \cite{paris2011local}. The image is from the under-exposure testing set \cite{wang2019underexposed}.}
\label{fig:supp_ours_vs_working_in_srgb}
\end{figure*}

\begin{table}[t]
\caption{Angular error of illuminant estimating using the image set captured by the Canon EOS-1Ds Mark III in the NUS dataset \cite{cheng2014illuminant}. We compare the results obtained by training a deep neural network on real raw-RGB training images, reconstructed (rec.) raw-RGB training images based on the standard XYZ reconstruction, and our CIE XYZ reconstruction. The best results are shown in bold.}\label{table:illuminant_estimation}
\scalebox{0.82}{
\begin{tabular}{|l|c|c|c|c|}
\hline

Training data & Mean & Median & Best 25\% & Worst 25\% \\ \hline
Real & 4.15 & 3.89 & 1.13 & 7.85  \\ \hline
Rec. (standard) & 3.37 & 3.03 & 1.05 & 6.68 \\ \hline
Real and rec. (standard) & 2.72 & 2.60 & 0.72 & 4.99 \\ \hline
Rec. (ours) & 3.00 & 2.61 & 0.83 & 5.37 \\ \hline
Real and rec. (ours) & \textbf{2.41} & \textbf{2.03} & \textbf{0.65} & \textbf{4.66} \\ \hline
\end{tabular}}
\end{table}

\begin{figure*}
\centering
\includegraphics[width=\linewidth]{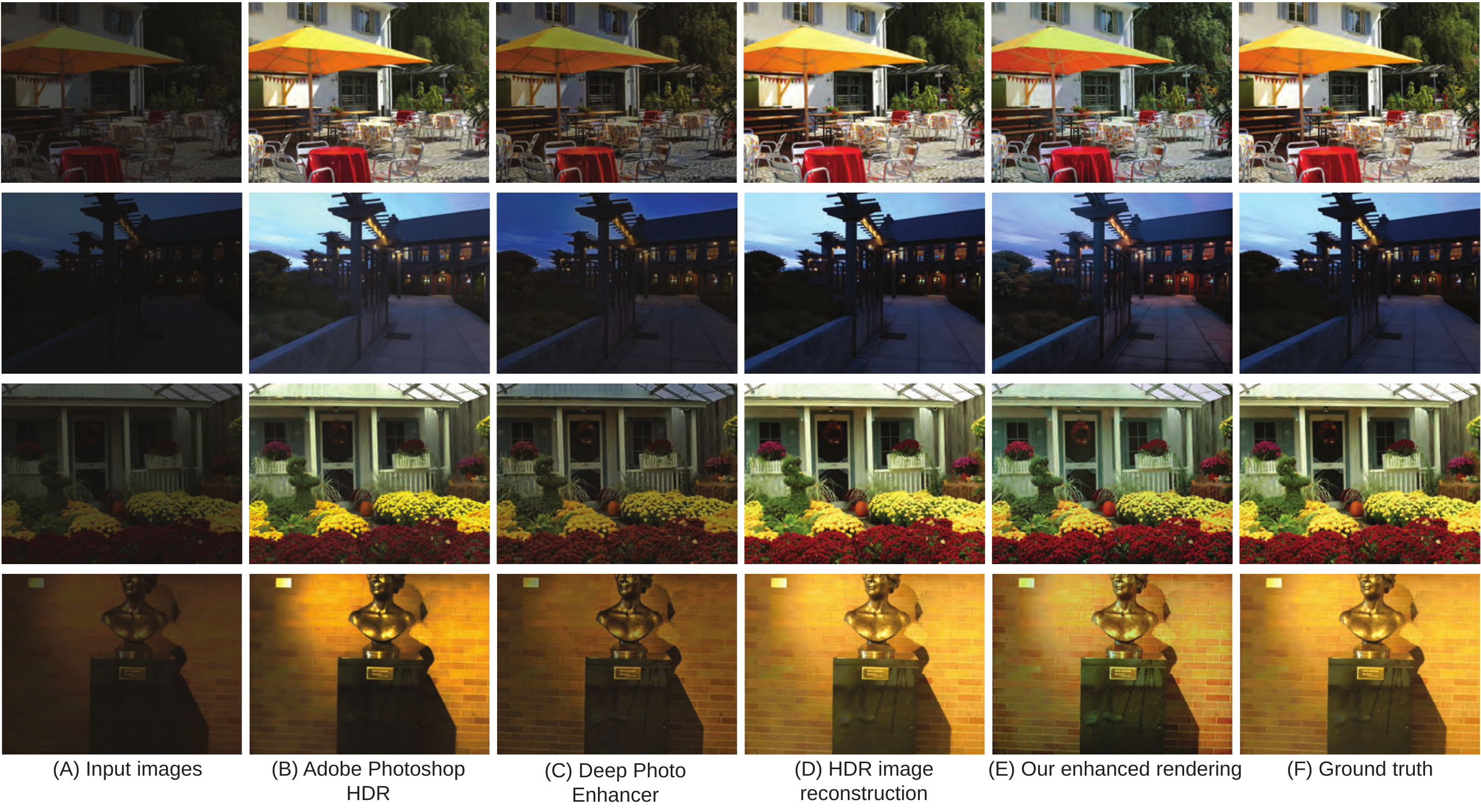}
\caption{Low-light image enhancement application. (A) Input sRGB rendered image. (B) Adobe Photoshop HDR results. (C) Deep Photo Enhancer results \cite{chen2018deep}. (D) HDR result of ~\cite{eilertsen2017hdr}. (E) Our re-rendered images after photo-finishing enhancement. (F) Ground truth images. Input images are taken from~\cite{eilertsen2017hdr}.}
\label{fig:supp_hdr}
\end{figure*}

\begin{figure*}
\centering\includegraphics[width=\linewidth]{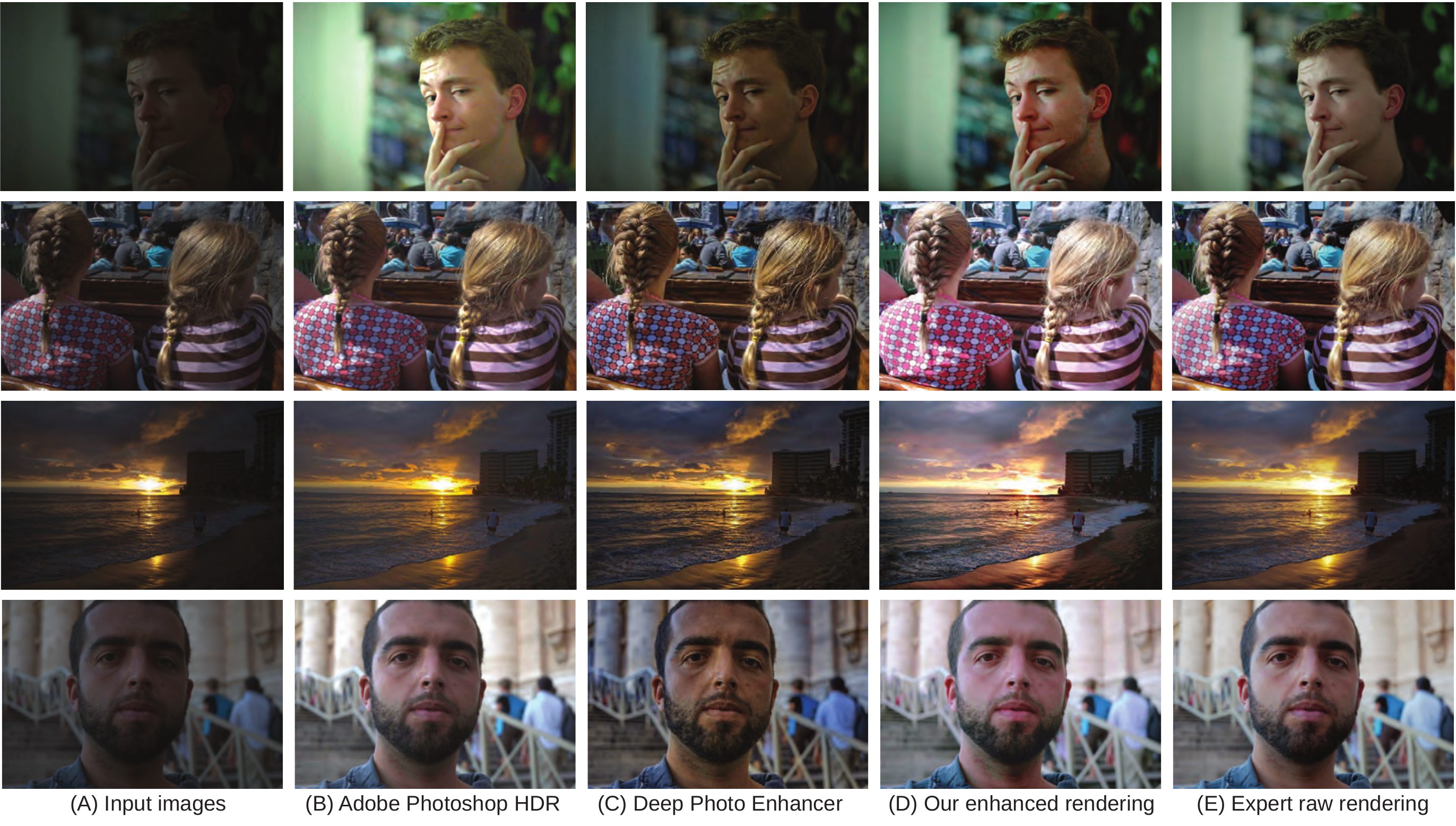}
\caption{Qualitative comparison for low-light image enhancement task. Images are taken from the under-exposure testing set \cite{wang2019underexposed}. (A) Input image. (B) Adobe Photoshop HDR results. (C) Results of deep photo enhancer \cite{chen2018deep}. (D) Our enhanced rendered image. (E) Expert-retouched image.}
\label{fig:supp_underexp}
\end{figure*}

We further evaluated this simple pipeline on 500 under-exposed images taken from \cite{wang2019underexposed}. Fig. \ref{fig:low-light} shows a qualitative example. We show a quantitative comparison in Table \ref{table:low-light}. %As shown, applying our simple operations obtains on par results with state-of-the-art deep learning methods that target low-light image enhancement.
Applying digital gain to our reconstructed space provides better results compared to using the standard XYZ reconstruction or the nonlinear sRGB space. This is due to the fact that our reconstructed images have a better linearization with a high tonal range; see Fig. \ref{fig:supp_ours_vs_working_in_srgb}. We provide additional results in Figs. \ref{fig:supp_hdr} and \ref{fig:supp_underexp}.

\begin{table}
	\scalebox{0.8}{
    \caption{Quantitative results of the photo-finishing enhancement application using 500 under-exposure images provided in \cite{wang2019underexposed}.\label{table:low-light}}

   \begin{tabular}{|c|c|} \hline
   \textbf{Method} & \textbf{PSNR} \\ \hline
     White-Box \cite{hu2018exposure}& 18.57 \\
     Distort-and-Recover \cite{park2018distort} & 20.97 \\
     HDRNet \cite{gharbi2017deep}& 21.96 \\
     Deep photo enhancer \cite{chen2018deep}& 22.150\\
     DeepUPE \cite{wang2019underexposed}& 23.04 \\ \hline
	 Enhanced in sRGB & 16.92  \\
     Enhanced in rec. standard XYZ & 18.41 \\\hdashline
	 Our enhanced re-rendering & 21.03 \\ \hline

     \end{tabular}}
\end{table}

\section{Concluding remarks}\label{sec:conclusion}

We have proposed a method and DNN model that can map back and forth between non-linear sRGB and linear CIE XYZ images more accurately compared to alternative approaches. Our method is based on learning a decomposition of sRGB images into a globally processed and locally processed image layers. The learned globally processed image layer is then used to learn a mapping to the device independent CIE XYZ color space.  We have provided extensive experiments targeting image restoration tasks including image denoising, deblurring, and defocus detection to show that our proposed model provides the performance boost anticipated from using linear images. In addition, utilizing the decomposed image layers, we show that our model can be used to perform various image enhancement and photo-finish tasks. Code and dataset are publicly available at \url{https://github.com/mahmoudnafifi/CIE_XYZ_NET}.

\appendix 
\section*{Additional Details} \label{additionaldetails}

\subsection*{Defocus Blur Detection}
\begin{figure*}
\centering
\includegraphics[width=\linewidth]{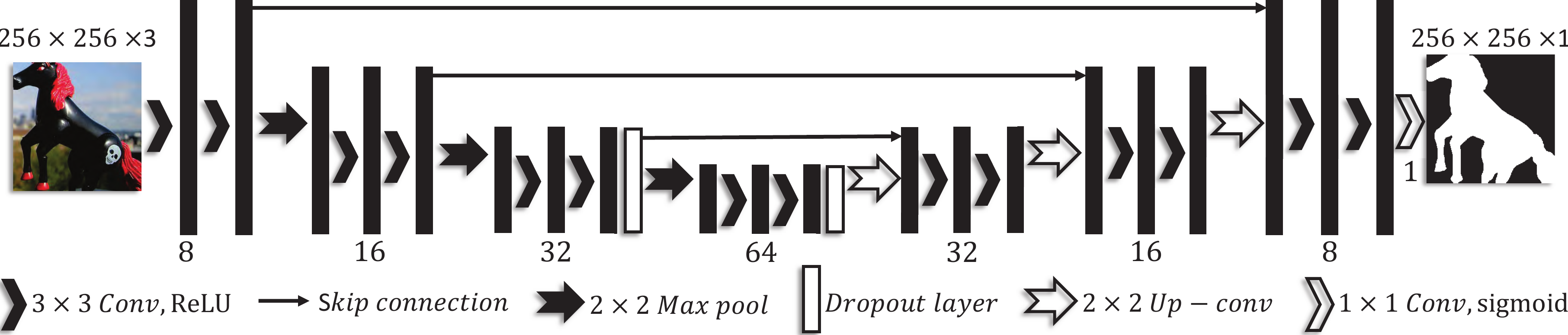}
\caption{We used a light-weight U-Net-like architecture for the task of defocus blur detection. The size of the input and output layers is shown above the image patches. The number of output filters is shown under the convolution operations.}
\label{fig:defocusUNet}
\end{figure*}

In Sec. \ref{secdefocus}, we used a light-weight U-Net-like architecture for the defocus blur detection task. Fig.~\ref{fig:defocusUNet} shows the details of the used architecture. We trained the model with image patches of size $256\times256$. We adopted He's weight initialization~\cite{he2015delving} and used the Adam optimizer~\cite{kingma2014adam} to train our model.

\begin{figure*}
\centering
\includegraphics[width=0.48\linewidth]{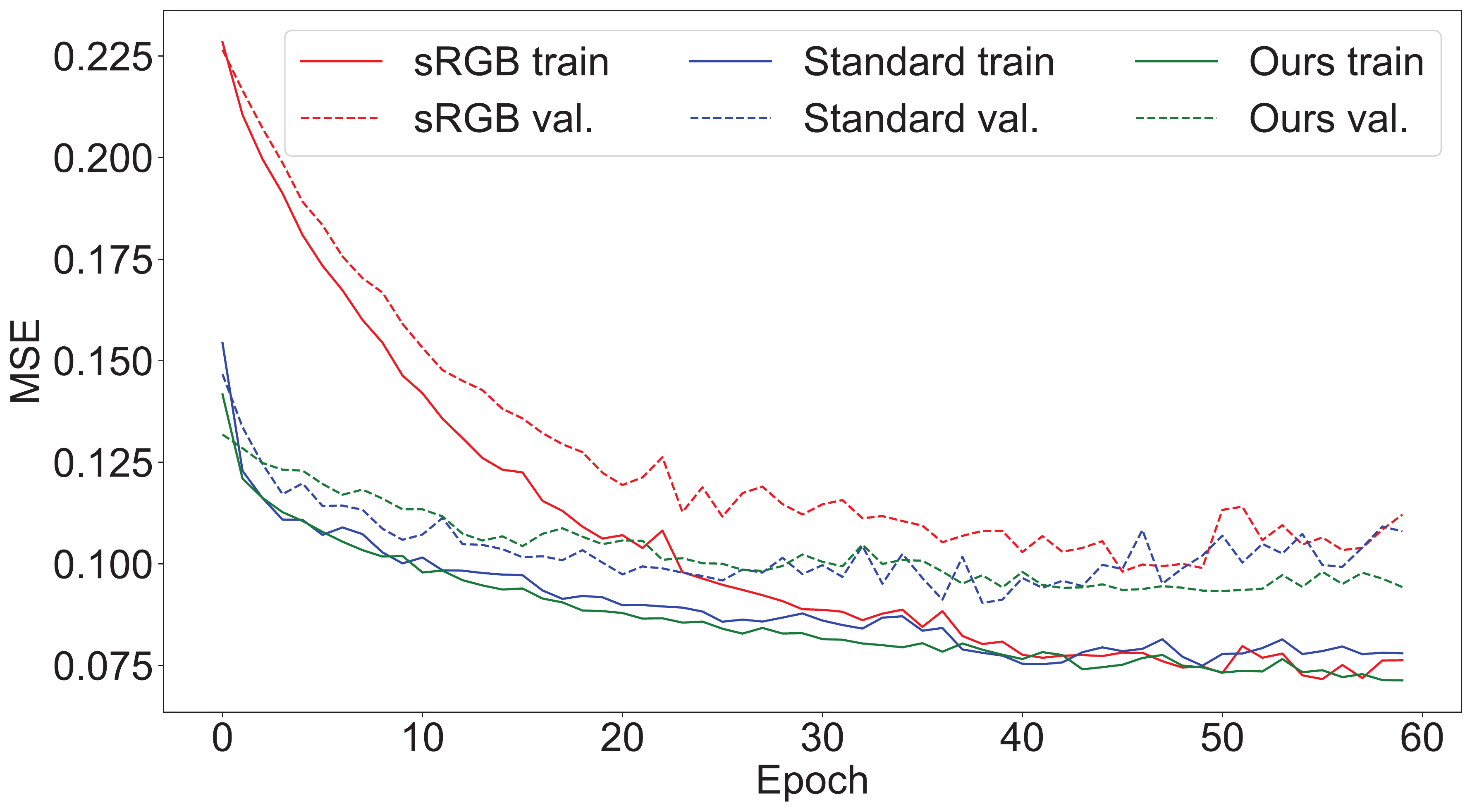}
\caption{Training and validation curves of different color spaces. The validation curve of our color space shows a better performance over time and achieves the smallest MSE compared to other curves.}
\label{fig:trainingCurves}
\end{figure*}

The initial learning rate is set to $10^{-4}$, which is decreased by half every 30 epochs. We train the model with mini-batches of size 12 using the mean squared error (MSE) loss between the output and the ground truth. During the training phase, we set the dropout rate to $0.5$. We found that the model converges after 60 epochs. In Fig.~\ref{fig:trainingCurves}, the training and validation curves of different color spaces are shown. Compared to other spaces, our reconstructed CIE XYZ space has a smooth validation convergence and is able to achieve faster convergence with the smallest MSE.

\subsection*{Raw-RGB Image Reconstruction}
We discussed the raw-RGB reconstruction as it is one of the potential applications of the proposed XYZ reconstruction in Sec. \ref{secrawrec}. In our experiments, we chose the scene illuminant estimation task to validate our raw-RGB reconstruction against the raw reconstruction based on the standard XYZ mapping~\cite{anderson1996proposal, ebner2007color}. We trained a deep model to estimate the scene illuminant from the given raw-RGB image. In this section, we provide the details of the model's architecture used in these experiments and the training details.

The model is designed to accept a $150\!\times\!150$ raw-RGB image (similar to prior work that proposed to use thumbnail images for the illuminant estimation task \cite{barron2017fast, afifi2019sensor}). The model includes a sequence of conv, leaky ReLU (LReLU), batch normalization (BN), and fully connected (FC) layers. In particular, the model consists of two conv--LReLU--conv--BN--LReLU blocks, followed by a conv--LReLU--FC--LReLU--dropout--FC--LReLU--FC block. All conv layers have $3\!\times\!3$ filters with a different number of output channels and stride steps. The first, second, and third conv layers have 64 output channels, while the fourth and fifth conv layers have 128 and 256 output channels, respectively. The stride steps were set to 2 for the first three conv layers. For the last two conv layers, we used a stride step of 3. The first two FC layers have 256 output neurons, while the last FC layer has 3 output neurons. We trained each model for 50 epochs to minimize the angular error between the estimated illuminant vector and the ground truth illuminant. The training process was performed with a learning rate of $10^{-4}$ and mini-batch of 32 using the Adam optimizer \cite{kingma2014adam} with a decay rate of gradient moving average 0.9 and a decay rate of squared gradient moving average 0.999.

\subsection*{Image Enhancement}
One of the potential applications of our method is low-light image enhancement. We exploited the higher tonal range in the reconstructed CIE XYZ image and proposed a simple set of heuristic operations to perform low-light image enhancement. In our experiments, we used the bilateral guided upsampling method \cite{chen2016bilateral} to speed up the running time required to apply the local details enhancement method~\cite{paris2011local} on the forward local sRGB reconstructed layer. Specifically, we apply the local details enhancement to a downsized version ($150\!\times\!150$) of the local layer, then we apply the bilateral guided upsampling to reconstruct the processed local layer in the original image's size.

\begin{figure*}
\centering
\includegraphics[width=\linewidth]{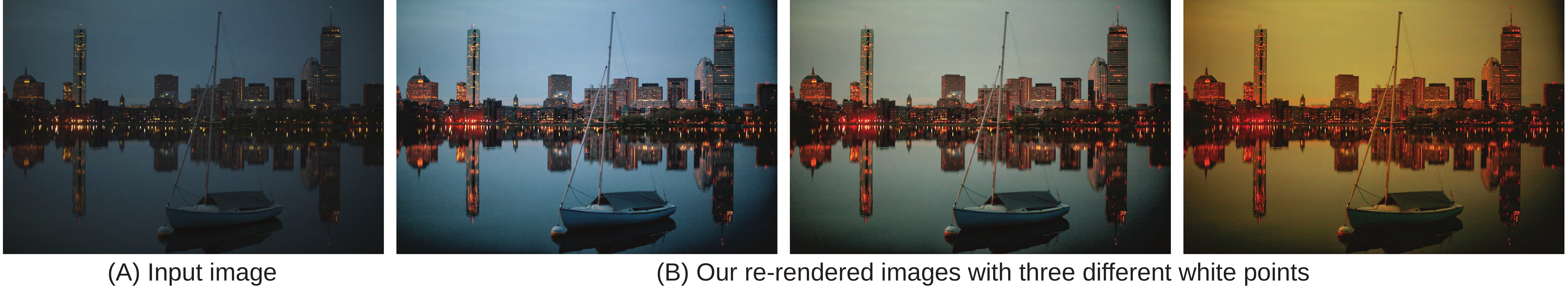}
\caption{(A) Input sRGB rendered image. (B) Our re-rendered images after enhancement. In this example, we applied chromatic adaptation to three different reference white points. Input image is taken from the under-exposure set \cite{wang2019underexposed} of the MIT-Adobe FiveK dataset \cite{fivek}.}
\label{fig:chromAdapt}
\end{figure*}

\section*{Additional Applications} \label{sec:applications}

When we work in our reconstructed space (i.e., XYZ), we have a sound interpretation of post-capture white-balance editing using standard white points (e.g., D65, D50) and standard chromatic adaptation transforms (e.g., Bradford CAT \cite{lam1985metamerism}, Sharp CAT \cite{finlayson1994spectral}), which are originally designed to work in the camera CIE XYZ space. Fig. \ref{fig:chromAdapt} shows examples of our enhanced rendering with applying chromatic adaptation \cite{finlayson1994spectral} in our reconstructed XYZ space.

\begin{figure*}
\centering
\includegraphics[width=0.95\linewidth]{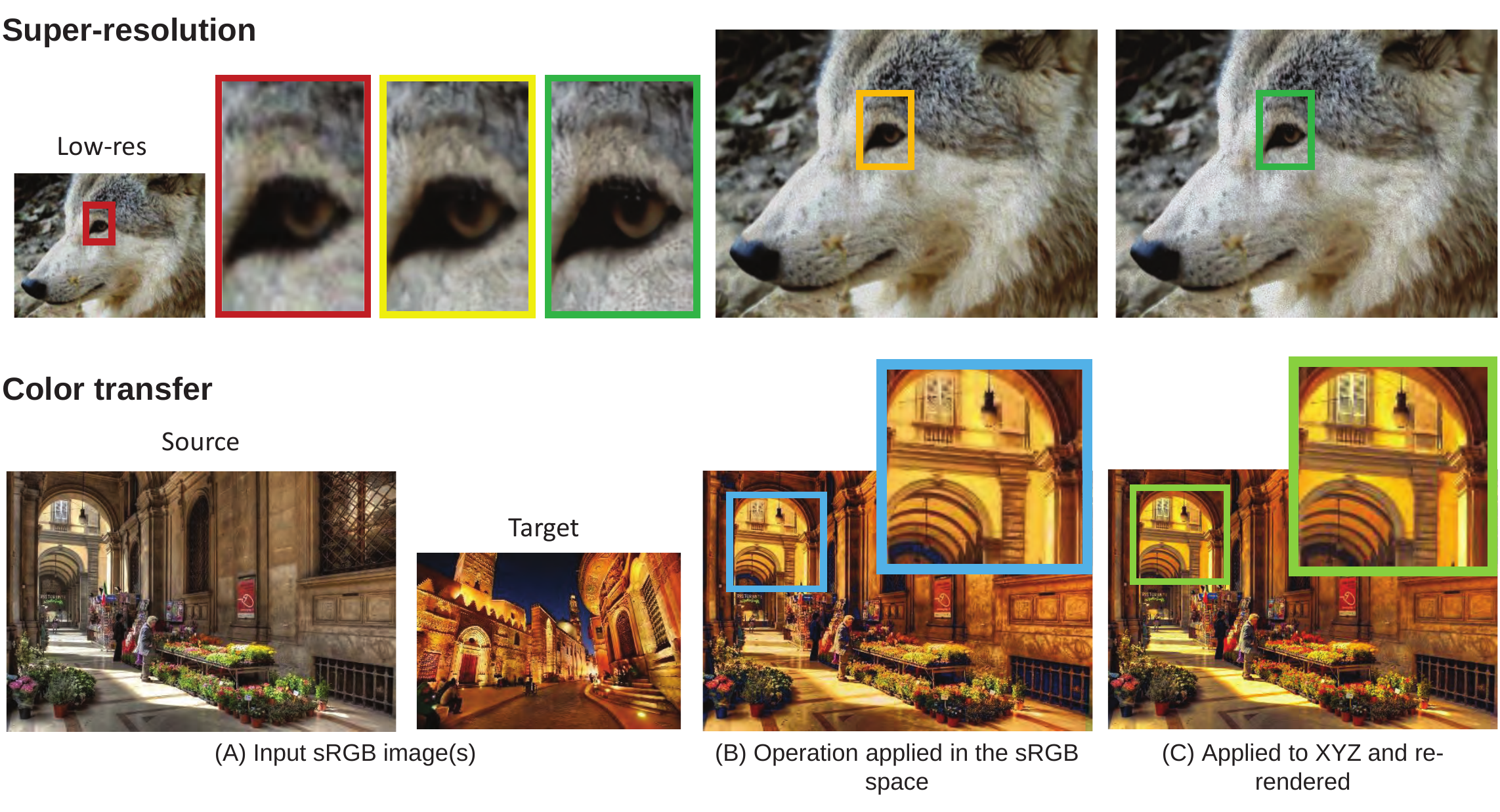}
\caption{Another potential application of our method. (A) The input sRGB image. (B) Super-resolution and color transfer applied in the sRGB space. (C) Super-resolution and color transfer applied in our reconstructed CIE XYZ space followed by re-rendering. In this example, we used the deep learning super-resolution model proposed in \cite{zhang2018learning} and the color transfer method in \cite{pitielinear}. The input image in the first row is taken from the DIV2K dataset \cite{Agustsson_2017_CVPR_Workshops, Timofte_2018_CVPR_Workshops}, while the second input image is taken from Flickr--CC BY-NC 2.0 (by Chris Ford and Giuseppe Moscato, respectively).}
\label{fig:otherapplication}
\end{figure*}

Additional potential applications are shown in Fig. \ref{fig:otherapplication}. In the first row of Fig. \ref{fig:otherapplication}, we show super-resolution results obtained directly by working in the sRGB space and in our reconstructed CIE XYZ space followed by applying our re-rendering process. The last row of Fig. \ref{fig:otherapplication} shows an arguably better color transfer result by applying the color transfer process in our reconstructed space.

\begin{figure*}
\centering
\includegraphics[width=\linewidth]{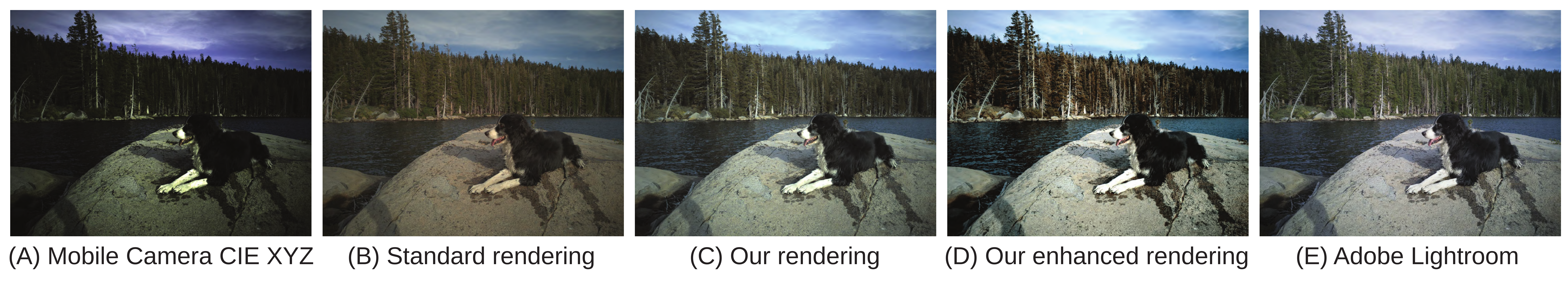}
\caption{Our rendering network generalizes well for unseen CIE XYZ input images and produces pleasing results that are close to Adobe Lightroom's quality. (A) Input smartphone camera CIE XYZ image. (B) Standard rendering \cite{anderson1996proposal, ebner2007color}. (C) Our rendering. (D) Our rendering after enhancing the local layer using the local Laplacian filter \cite{paris2011local}. (E) Adobe Lightroom rendering. To aid visualization, CIE XYZ images are scaled by a factor of two. Input image is taken from the HDR+ burst photography dataset \cite{hasinoff2016burst}.}
\label{fig:mobileCIE-XYZ}
\end{figure*}

Lastly, our rendering network can be used as an alternative way to produce aesthetic photographs from raw-RGB DNG files, as shown in Fig. \ref{fig:mobileCIE-XYZ}. In this example, we first used the DNG metadata to map the raw-RGB values into the CIE XYZ space. Then, we used our rendering network and a local Laplacian filter to generate the shown output images.

\end{document}